\documentclass[sigconf]{acmart}

\AtBeginDocument{%
  \providecommand\BibTeX{{%
    \normalfont B\kern-0.5em{\scshape i\kern-0.25em b}\kern-0.8em\TeX}}}

\copyrightyear{2024}
\acmYear{2024}
\setcopyright{acmlicensed}\acmConference[MM '24]{Proceedings of the 32nd
ACM International Conference on Multimedia}{October 28-November 1,
2024}{Melbourne, VIC, Australia}
\acmBooktitle{Proceedings of the 32nd ACM International Conference on
Multimedia (MM '24), October 28-November 1, 2024, Melbourne, VIC, Australia}
\acmDOI{10.1145/3664647.3680723}
\acmISBN{979-8-4007-0686-8/24/10}

\usepackage[utf8]{inputenc} 
\usepackage[T1]{fontenc}    
\usepackage{hyperref}       
\usepackage{url}            
\usepackage{booktabs}       
\usepackage{amsfonts}       
\usepackage{nicefrac}       
\usepackage{microtype}      
\usepackage{xcolor}         
\usepackage{graphicx}

\usepackage{float}
\usepackage{subfigure}
\usepackage{array} 
\usepackage{threeparttable}
 
\usepackage{amsmath,amssymb,latexsym}
\usepackage{caption}   
\usepackage{overpic} 
\usepackage{multicol}
\usepackage{multirow}
\usepackage{makecell}
\usepackage{wrapfig}
\usepackage{makecell}
\usepackage{microtype}
\usepackage{balance}
\usepackage{enumitem}

\begin{document}

\title{Multi-Granularity Hand Action Detection}


\author{Ting Zhe}
\affiliation{%
  \institution{School of Computer Science \\Wuhan University}
  \city{Wuhan}
  \country{China}}
\email{zheting@whu.edu.cn}

\author{Jing Zhang}
\affiliation{%
  \institution{The University of Sydney}
  \city{Sydney}
  \country{Australia}}
\email{jingzhang.cv@gmail.com}

\author{Yongqian Li}
\affiliation{%
  \institution{School of Computer Science \\Wuhan University}
  \city{Wuhan}
  \country{China}}
\email{yongqianli@whu.edu.cn}

 \author{Yong Luo$^{\dag}$}
 \affiliation{
   \institution{Wuhan University \&  \\Hubei Luojia Laboratory}
   \city{Wuhan}
   \country{China}}
 \email{luoyong@whu.edu.cn}
 
 \author{Han Hu}
 \affiliation{
   \institution{Beijing Institute of Technology}
   \city{Beijing}
   \country{China}}
 \email{hhu@bit.edu.cn} 

 \author{Dacheng Tao}
 \affiliation{%
  \institution{Nanyang Technological University}
  \city{Singapore}
  \country{Singapore}}
\email{dacheng.tao@ntu.edu.sg}

\thanks{\dag Corresponding author: Yong Luo}
\renewcommand{\shortauthors}{Ting Zhe et al.}

\begin{abstract}
 Detecting hand actions in videos is crucial for understanding video content and has diverse real-world applications. Existing approaches often focus on whole-body actions or coarse-grained action categories, lacking fine-grained hand-action localization information. To fill this gap, we introduce the \textbf{FHA-Kitchens} (Fine-Grained Hand Actions in Kitchen Scenes) dataset, providing both coarse- and fine-grained hand action categories along with localization annotations. This dataset comprises 2,377 video clips and 30,047 frames, annotated with approximately 200k bounding boxes and 880 action categories. Evaluation of existing action detection methods on FHA-Kitchens reveals varying generalization capabilities across different granularities. To handle multi-granularity in hand actions, we propose \textbf{MG-HAD}, an End-to-End \textbf{M}ulti-\textbf{G}ranularity \textbf{H}and \textbf{A}ction \textbf{D}etection method. It incorporates two new designs: Multi-dimensional Action Queries and Coarse-Fine Contrastive Denoising. Extensive experiments demonstrate MG-HAD's effectiveness for multi-granularity hand action detection, highlighting the significance of FHA-Kitchens for future research and real-world applications. The dataset and source code are available at \href{https://github.com/superZ678/MG-HAD}{MG-HAD}.
\end{abstract}





\maketitle

\begin{figure}[t!]
  \centering
\includegraphics[width=0.48\textwidth,height=0.32\textwidth]{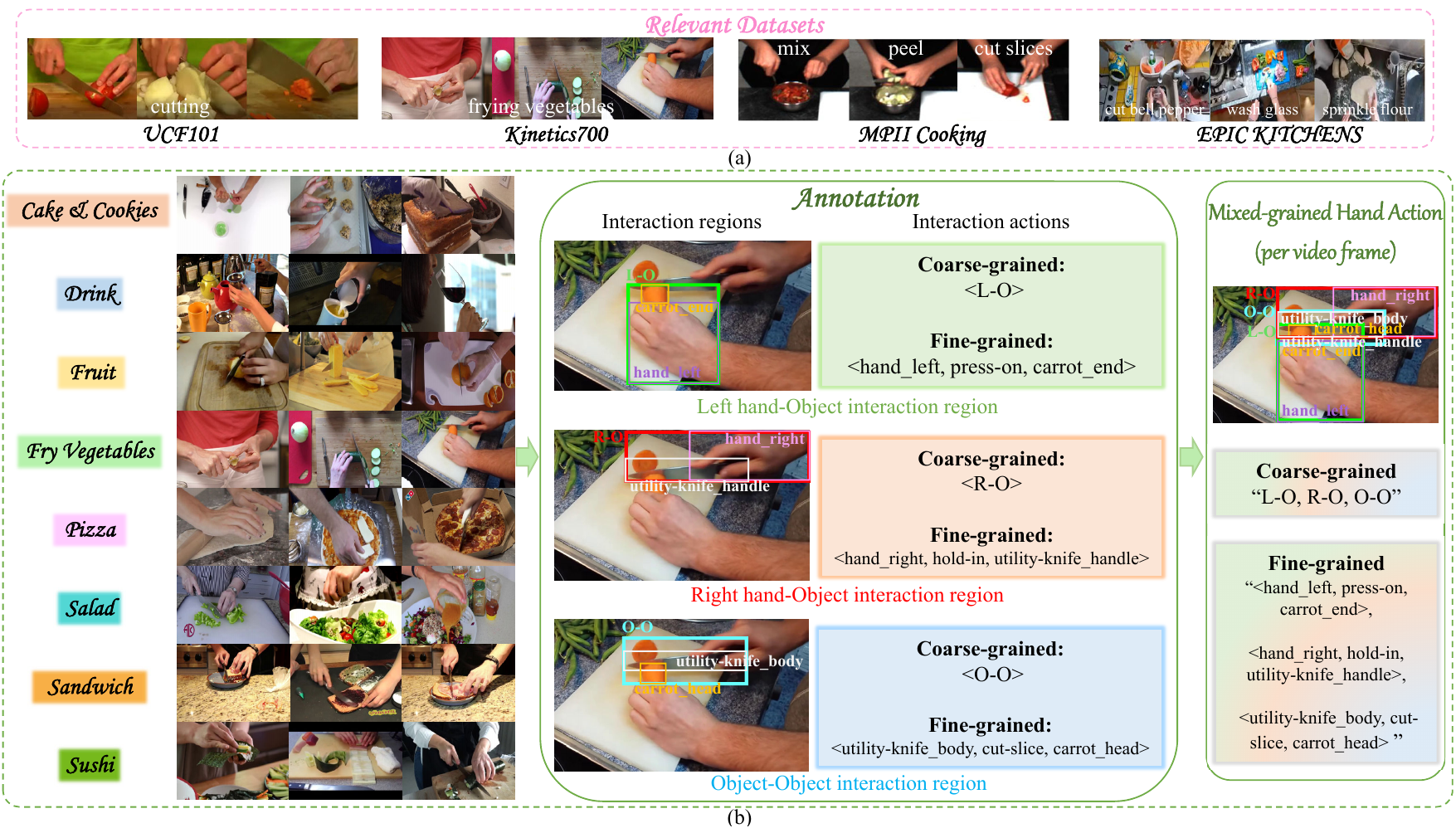}
  \caption{Overview of the \emph{\textbf{FHA-Kitchens}} dataset. (a) The annotation of hand actions in existing relevant datasets, where UCF101~\cite{3ucf101} and Kinetics700~\cite{700_2020} are whole-body action datasets, while MPII Cooking~\cite{2MPII} and EPIC KITCHENS~\cite{1epic} are hand action datasets. (b) The annotation of hand actions in our dataset. The left shows some frames extracted from 8 dish categories. The right illustrates the annotation process of hand actions in \emph{``fry vegetable''}.} \label{fig:openfig}
\end{figure}

\section{Introduction}
 Action detection, a crucial task in video understanding, aims to locate and recognize action instances in each video frame, with applications in various fields~\cite{zhang2020empowering} such as Human-Computer Interaction (HCI)~\cite{49hand-object}, Smart Homes~\cite{55behavior}, the Design and Control of Robot Hands~\cite{54robothand}, and Healthcare~\cite{48RCareWorld}. Despite significant advancements in action recognition regarding both large-scale benchmarks~\cite{3ucf101,6kinetic400} and advanced algorithms~\cite{19TSN,22slowfast,23videoswin}, action detection remains relatively underexplored, mainly due to the lack of datasets with spatial action localization annotations. Moreover, existing methods predominantly focus on whole-body actions, overlooking the fine-grained actions of specific body parts, such as hands. However, hand actions are integral to daily activities, underscoring the significant research and practical importance of hand action detection.

 Pioneering datasets such as MPII Cooking Activities~\cite{2MPII} and EPIC-KITCHENS~\cite{1epic} have been developed to facilitate hand-action research. However, they exhibit limitations including insufficient representation of hand-action granularity, lack of annotation of hand-action interaction regions, and neglect of the relationships between interacting objects. As shown in Figure~\ref{fig:openfig}(a), they offer only coarse-grained annotations for hand actions like ``\emph{cut}'' rather than the fine-grained multi-dimensional categories like ``\emph{<knife, cut slice, carrot>}''. These limitations hinder the study of detecting fine-grained hand actions and exploring their spatial relationship, leaving challenges in hand action detection unresolved. Therefore, establishing a large-scale benchmark with rich hand-action annotations is essential for advancing research in this field.

 To this end, this paper presents a novel dataset \textbf{FHA-Kitchens}, focusing on rich and fine-grained localization and categorization information of hand actions in kitchen scenes. The FHA-Kithcens dataset encompasses a total of 2,377 video clips and 30,047 frames from eight different dish types (Figure~\ref{fig:openfig}(b) left). Each frame includes meticulously annotated hand action information, featuring high-quality annotations of hand interaction region boxes and corresponding coarse- and fine-grained categories. Our data were extracted from publicly available large-scale action datasets~\cite{700_2020}, focusing on videos relevant to hand actions. Subsequently, frames underwent cleaning and were annotated by ten expert voluntary annotators. To excavate more hand action information, we refined the annotation process in two aspects (Figure~\ref{fig:openfig}(b) right): (1) \textbf{Hand interaction regions}. These were subdivided into three sub-regions based on hand-object interaction: Left hand-Object interaction region, Right hand-Object interaction region, and Object-Object interaction region. Each sub-interaction region was annotated with bounding boxes and coarse-grained categories, denoted as ``L-O'', ``R-O'', and ``O-O''. (2) \textbf{Hand interaction actions}. To enhance the model's understanding of hand actions, we further refined the category of each sub-interaction region by expanding single-dimensional action categories into multi-dimensional ones, annotated in the format of triplets: \emph{<subject, action verb, object>}, abbreviated as \emph{<s, a, o>}, where ``s-o'' denotes interacting objects, and ``a'' represents the interaction action between the objects. Additionally, when annotating ``s'' and ``o'', we considered the specific contact area between the interacting objects, labeled as ``name\_contact area'' (\textit{e.g.}, ``\emph{carrot\_end}''). Overall, we meticulously annotated \textbf{880} hand action categories (coarse- and fine-grained) for approximately 220k bounding boxes, with each category corresponding to a sub-interaction region's localization box. Fine-grained categories per frame have nine dimensions, resulting in 877 action triplets.

Hand Action Detection (HAD) is a sub-area of Action Detection (AD) research, which has a close relation to generic object detection (OD) in the image domain. We systematically evaluated several representative AD methods~\cite{24faster,50yolox,defordetr,dino} on FHA-Kitchens, observing varied performance across different levels of granularity: ``Coarse-grained'' and ``Fine-grained''. Existing methods perform significantly worse under fine-grained labels compared to coarse-grained ones, indicating that these detection methods have a better understanding of single-dimensional coarse-grained labels (\textit{i.e.}, single verb or noun). However, real-world hand actions often involve both coarse- and fine-grained information simultaneously. Therefore, exploring the impact of multi-granularity action categories in HAD tasks is both interesting and practically significant.

\begin{table*}[htbp]
  \caption{Comparison of relevant datasets. AR: Action Recognition. AD: Action Detection. HAD: Hand Action Detection. OD: Object Detection. ACat.: Action Category. OCat.: Object Category. Dim: Action Dimension. IRBox: Interaction Region Box.} 
\label{tab:datasetcomparison}
\centering
\begin{threeparttable}
\scalebox{0.85}{
\begin{tabular}{cp{0.5cm}<{\centering}p{0.3cm}<{\centering}p{1.1cm}<{\centering}p{1.43cm}<{\centering}p{0.55cm}<{\centering}p{1.5cm}<{\centering}p{0.68cm}<{\centering}cp{1.2cm}<{\centering}p{1cm}<{\centering}p{2.5cm}<{\centering}} 
\toprule 
Dataset & Year & {Ego} & \#Clip & {Ave.Len} & {\#Frame} & \#ACat. & \#Verb & \#OCat. & Dim & IRBox & Task\\
\midrule
\textbf{Whole-body action dataset} &&&&&&&&&&& \\ 
\hline
UCF101~\cite{3ucf101} & 2012 & $\times$ & 13.3K & $\sim$6\emph{s} & - & 101 & - & - & 1 & $\times$ & AR \\ 
ActivityNet~\cite{5activitynet} & 2015 & $\times$ & 28K & [5,10]\emph{m} & - & 203 & - & - & 1 & $\times$ & AR \\ 
Kinetics400~\cite{6kinetic400} & 2017 & $\times$ & 306K & 10\emph{s} & - & 400 & 359 & 318 & 2 & $\times$ & AR \\ 
Kinetics600~\cite{7kinetics600} & 2018 & $\times$ & 496K & 10\emph{s} & - & 600 & 550 & 502 & 2 & $\times$ & AR \\ 
Kinetics700~\cite{8kinetic700} & 2019 & $\times$ & 650K & 10\emph{s} & - & 700 & 644 & 591 & 2 & $\times$ & AR \\ 
AVA~\cite{10ava} & 2018 & $\times$ & 430 & 15\emph{m} & - & 80 & 80 & 0 & 3 &  $\times$  & AR,AD \\ 
AVA-kinetics~\cite{11ava-kinetics} & 2020 & $\times$ & 230K & 15\emph{m},10\emph{s} & - & 80 & 80 & 0 & 3 &  $\times$  & AR,AD \\ 
FineGym~\cite{9finegym} & 2020 & $\times$ & 32K & 10\emph{m} & - & 530 & 530 & 0 & 3 & $\times$ & AR \\  
\hline
\textbf{Hand action dataset} &&&&&&&&&&& \\ 
\hline
MPII cooking~\cite{2MPII} & 2012 & $\times$ & 5,609 & 15\emph{m} & 881K & 65 & 65 & 0 & 1 & $\times$ & AR \\ 
EPIC-KITCHENS~\cite{1epic} & 2018 & \checkmark & 39.6K & 3.7$\pm$5.6\emph{s} & 11.5M & 149 & 125 & 323 & 2 & $\times$ & AR,OD \\ 
\textbf{FHA-Kitchens} & \textbf{2024} & \textbf{\checkmark} &\textbf{2,377} & \textbf{3\emph{m}} &\textbf{30,047} & \textbf{880} & \textbf{130} & \textbf{384} & \textbf{9}  & \textbf{\checkmark} & \textbf{AR,AD,HAD,OD} \\
\bottomrule
 \end{tabular} }
\end{threeparttable}
\end{table*}

 Among the state-of-the-art detection methods, DINO~\cite{dino} showed relatively strong performance across different granularity hand actions. Building upon DINO, we propose \textbf{MG-HAD}, a novel baseline for hand action detection (HAD). MG-HAD contains a backbone, a multi-layer Transformer encoder, a multi-layer Transformer decoder, and multiple prediction branches. To better adapt to multi-granularity hand actions, we propose two novel designs: (1) \textbf{Multi-dimensional information processing}: To enhance the model's understanding of fine-grained information, we replace the original single-dimensional content query in the decoder with multi-dimensional content queries to focus on multiple aspects of hand actions. Additionally, we introduce a Content Query Reorganization (CQR) module to generate three query sets focusing on different action dimensions as decoder inputs. (2) \textbf{Multi-granularity category processing}: We observed that the DINO's CDN (Contrastive DeNoising) module mainly focuses on bounding boxes for contrastive denoising training, while the labels are not specially designed. To enable the model to better learn and distinguish coarse-grained and fine-grained action labels, we devise coarse-grained and fine-grained sample queries for contrastive denoising training of the labels, by adding noise to different granularity categories with specified noise positions and classes. Besides, we investigate the pre-trained ResNet50~\cite{restnet} and Swin-L~\cite{swin} models as backbones to extract multi-scale visual features. During training, following the DN-DETR~\cite{dndetr} method, we add ground truth labels and boxes with noises into the Transformer decoder layers to stabilize bidirectional matching, and also adopt deformable attention~\cite{defordetr} for improved computational efficiency. 

 In summary, our contributions can be summarized as follows:
 
 \begin{itemize}[leftmargin=*]
 \item To the best of our knowledge, we are the first to study the problem of multi-granularity hand action detection and establish the first hand-action dataset \textbf{FHA-Kitchens}, which includes both hand interaction region localization and multi-granularity category annotations. This dataset can serve as a benchmark for hand action detection tasks. 
 \item We systematically investigated the impact of different granularity hand action information in kitchen scenes on the hand action detection task and provided insights about the evaluation protocol, performance analysis, and model design. 
 \item We propose a novel multi-granularity hand action detection method named \textbf{MG-HAD}, which is designed from the perspectives of multi-granularity and multi-dimensionality. This method incorporates Multi-dimensional Action Queries and a Coarse-Fine Contrastive Denoising module to address the mixed-grained HAD problem. MG-HAD demonstrates its effectiveness in hand action detection and could serve as a strong baseline.
\end{itemize}

\section{Related work}

\subsection{AR \& AD Dataset}
\textbf{Action Recognition (AR) Dataset.} 
Existing studies on action recognition datasets can be divided into two main categories based on the types of actions: whole-body action and part-body action, such as UCF101~\cite{3ucf101}, Kinetics~\cite{6kinetic400,7kinetics600,8kinetic700}, ActivityNet~\cite{5activitynet}, FineGym~\cite{9finegym}, and others~\cite{12moments,13,15,14,16,35KTH,4Weizmann}. These datasets primarily focus on whole-body actions, lacking fine-grained action information from specific body parts. Datasets like MPII Cooking Activities~\cite{2MPII} and EPIC-KITCHENS~\cite{1epic} refine the action verb part and consider interacting objects, yet they do not describe the localization of action interaction regions or the relationships between interacting objects, crucial for HAD tasks.

\textbf{Action Detection (AD) Dataset.} Compared to action recognition datasets, fewer datasets are available for action detection~\cite{10ava,11ava-kinetics}. This is due to the need to annotate the position and category of each action instance, which requires more effort for dataset construction. The AVA dataset~\cite{10ava} focuses on human action localization, providing bounding box annotations for each person. However, this dataset primarily focuses on whole-body actions, providing location information for individuals rather than action interaction regions. Moreover, the provided action categories are mainly single-dimensional coarse-grained verbs (\textit{e.g.}, ``\emph{sit}'', ``\emph{write}'', and ``\emph{stand}''). FHA-Kitchens dataset addresses these limitations by providing precise bounding box annotations for each hand sub-interaction region. A comprehensive comparison between FHA-kitchens and existing datasets is presented in Table~\ref{tab:datasetcomparison}. In contrast to existing datasets, \textbf{(1)} We provide precise localization information by meticulously annotating hand interaction regions and corresponding interaction objects using bounding boxes. \textbf{(2)} We offer two granularity for hand actions: coarse- and fine-grained. For fine-grained categories, we use multi-dimensional triplets to represent each sub-interaction region action, expanding the dimensionality of each frame to 9. \textbf{(3)} We not only focus on the interacting objects that generate interaction actions but also consider the active and passive relationships between these objects, capturing their contact areas.

\subsection{AR \& AD Method}
\textbf{Action Recognition (AR) Method.} Existing action recognition methods can be broadly summarized into two pipelines based on technical approaches. The first pipeline employs a 2D CNN~\cite{32,33,36,40} to learn frame-level semantics and then aggregate them temporally using 1D modules. For example, TSN~\cite{19TSN} divides an action instance into multiple segments, represents it with a sparse sampling scheme, and applies average pooling to fuse predictions from each frame. TRN~\cite{34} and TSM~\cite{37tsm} replace pooling with temporal reasoning and shift modules, respectively. The second pipeline directly utilizes a 3D CNN~\cite{6kinetic400,21non-local,22slowfast,23videoswin,38,41} to capture spatial-temporal semantics, such as I3D~\cite{6kinetic400}, SlowFast~\cite{22slowfast}, and Video Swin Transformer~\cite{23videoswin}. On the other hand, AR methods can be categorized into coarse-grained~\cite{30,31} and fine-grained~\cite{26,27,28,29} based on the granularity of the actions.

\textbf{Action Detection (AD) Method.} Most state-of-the-art action detection methods~\cite{cycleacr,x3d,22slowfast,46,35} commonly follow a two-stage pipeline, utilizing separate 2D and 3D backbones for localization and video feature extraction, respectively. Since transformer~\cite{attention} was introduced for machine translation, it has become a widely adopted backbone for sequence-to-sequence tasks~\cite{Icaption,dino,videomaev2,vitpose,vitpose++}. Most recent methods~\cite{vat,react,evad,7,tuber,stmixer} utilize a unified backbone to perform action detection. VAT~\cite{vat} is a transformer-style action detector designed to aggregate spatiotemporal context around target actors. EVAD~\cite{evad}, built upon the ViT framework, offers an end-to-end efficient video action detection method. WOO~\cite{7} and TubeR~\cite{tuber} are query-based action detectors that follow the detection frameworks of~\cite{sparse,detr} to predict bounding boxes and action classes, while STMixer~\cite{stmixer} is a one-stage query-based detector that adaptively samples discriminative features. However, we observed that these methods primarily focus on individual human actions and overlook action interaction regions, interacting objects, and their relationships. Leveraging the advantages of transformer-based detection models, we propose an end-to-end solution capable of simultaneous hand action localization and recognition.

\section{FHA-Kitchens Dataset}

\subsection{Data Collection And Organization}
\textbf{Data Collection.} 
The proposed dataset is derived from the large-scale action dataset Kinetics 700\_2020~\cite{700_2020}, which comprises approximately 650K YouTube video clips and over 700 action categories. To narrow our focus to hand actions, we performed filtering and processing operations on the original videos in three steps. \textbf{(1)} Content Localization: We observed that videos in kitchen scenes prominently showcase human hands. So we sought out and extracted relevant videos set against a kitchen backdrop. \textbf{(2)} Quality Selection: To ensure dataset quality, we selectively chose videos with higher resolutions. Specifically, 87\% of the videos were recorded at 1,280 $\times$ 720 resolution, while another 13\% had a shorter side of 480. Additionally, 67\% of the videos were captured at 30 frames per second (fps), and another 33\% were recorded at 24$\sim$25 fps. \textbf{(3)} Duration Control: We imposed a duration constraint on the videos, ranging from 30 seconds to 5 minutes, to exclude excessively long videos. This constraint aimed to maintain a balanced distribution within the sample space. Finally, we collected a total of 2,377 video clips, amounting to 84.22 minutes of footage, encompassing 8 distinct types of dishes.

\textbf{Data Organization.} 
The collected video data was reorganized and cleaned to align with our annotation criteria (Section~\ref{3.2}). At last, we obtained a total of 30,047 high-quality candidate video frames containing diverse hand actions for the FHA-Kitchens dataset. Compared to the initial collection, 113,436 frames were discarded during the cleaning process.

\subsection{Data Annotation}~\label{3.2}
We recruited 10 voluntary annotators to annotate hand actions for each frame with high quality, including bounding boxes and action categories. The annotation content and criteria are detailed below.
 
\textbf{Bounding Box Annotation:}  We annotated the bounding boxes for both interaction regions (IR) and interaction objects (IO). 
(1) \textbf{IR}: We divided the hand's interaction region into three sub-interaction regions: Left hand-Object (L-O), Right hand-Object (R-O), and Object-Object (O-O) interaction regions (Figure~\ref{fig:openfig}(b) middle), representing regions where the left hand directly contacts an object, the right hand directly contacts an object, and objects interact with each other, respectively. The reason for focusing on O-O is that interactions between objects also involve the participation of hands.
(2) \textbf{IO}: To better understand interaction actions, we also annotated the interactive object pair within each sub-interaction region using bounding boxes. For example, in L-O, we annotated objects directly touched by the left hand. In O-O, we annotated the interacting objects directly involved in hand actions (\textit{e.g.}, \textit{utility knife}  and \textit{carrot}). However, during annotation, we may encounter overlapping bounding boxes, \textit{i.e.}, the same interacting object will satisfy two annotation definitions, for example, the \textit{utility knife} in Figure~\ref{fig:openfig}, which is both the object directly touched by the right hand in the R-O and the active force provider in the O-O. In this case, we annotate all the labels because the same object participates in different interaction actions and has different roles (Annotation details can be seen in \emph{supplementary material}). Finally, we annotated a total of 198,839 bounding boxes, including 49,746 hand boxes, 66,402 interaction region boxes, and 82,691 interaction object boxes.  

\textbf{Hand Action Annotation:} We annotated coarse- and fine-grained action categories for each sub-interaction region. Coarse-grained categories, denoted by the generic terms ``L-O'', ``R-O'', and ``O-O'', represent the coarse actions within the sub-interaction regions. Different from existing fine-grained datasets. We annotate each fine-grained action category in a triplet format: \emph{\textless subject, action verb, object\textgreater}.
(1) \textbf{Subject $\&$ Object}: We considered the \emph{``active-passive''} relationship between objects, where the ``subject '' refers to the active force provider (\textit{e.g.}, \textit{utility knife}) and the ``object'' refers to the passive force receiver (\textit{e.g.}, \textit{carrot}), and annotate them in order within the action triplet. In L-O or R-O, the subject represents the corresponding hand, while the object denotes the directly interacting object. Furthermore, to enrich the description of each action, we also considered the contact areas of interacting objects within each sub-interaction region. For example, as shown in the first green block in the middle of Figure~\ref{fig:openfig}(b), we labeled the subject as ``\emph{hand\_left}'' and the object as ``\emph{carrot\_end}''. We referred to the EPIC-KITCHENS~\cite{1epic} dataset to define the object noun. (2) \textbf{Action Verb}: It describes the fine-grained hand action within the sub-interaction region. We used fine-grained verbs in the annotated action triplets and constructed the verb vocabulary by sourcing from EPIC-KITCHENS~\cite{1epic}, AVA~\cite{10ava}, and Kinetics 700~\cite{8kinetic700}.

\begin{figure}[t!]
 \centering
  \centering
  \includegraphics[width=7.2cm]{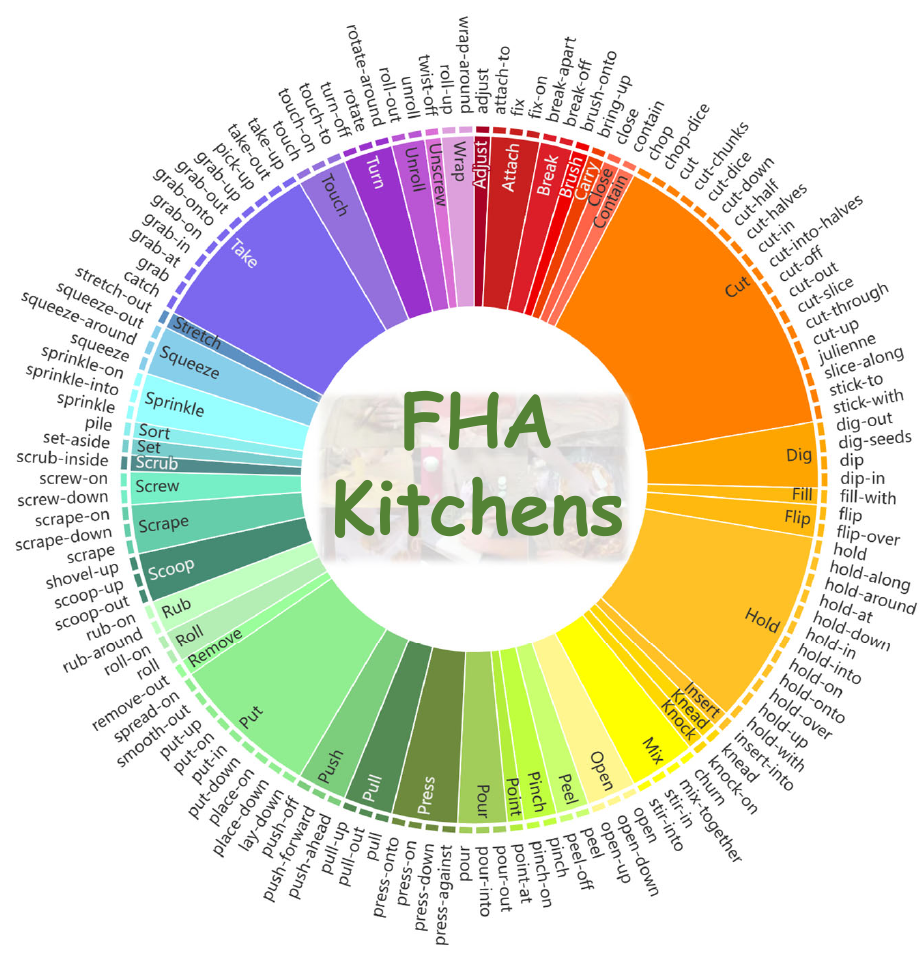}
  \caption{An overview of the action verbs and their parent action categories in FHA-Kitchens.} \label{fig:cakefig-actionverbs}
\end{figure}

\subsection{Statistics of the FHA-Kitchens Dataset}
\textbf{Overview of FHA-Kitchens.} As summarized in Table~\ref{tab:datasetcomparison}, we annotated hand action information for 30,047 frames from 2,377 clips, resulting in 880 action categories (including 877 action triplets), 130 action verbs, and 384 interaction object nouns. We have taken steps to refine the dataset by focusing on hand action categories and interaction regions, providing more precise localization bounding boxes and rich hand action categories for the three sub-interaction regions. Compared to the original action annotations in Kinetics 700\_2020~\cite{700_2020}, the FHA-Kitchens dataset expands the action labels by 7 dimensions, increases the number of action categories by 52 times, and introduces 122 new action verbs. Furthermore, we provide bounding boxes for hand action regions (\textit{i.e.}, 66,402 interaction region boxes). This expansion significantly enhances the diversity of hand action annotations, provides valuable region-level contextual information for each action, and facilitates future research for a wider range of video understanding tasks. The FHA-Kitchens dataset is then randomly divided into the disjoint train, validation, and test sets, with a video clip-based ratio of 7:1:2.

\begin{figure}[htbp]
 \centering
 \begin{minipage}[!htb]{0.48\textwidth}
  \centering
  \includegraphics[width=8.3cm,height=2.8cm]{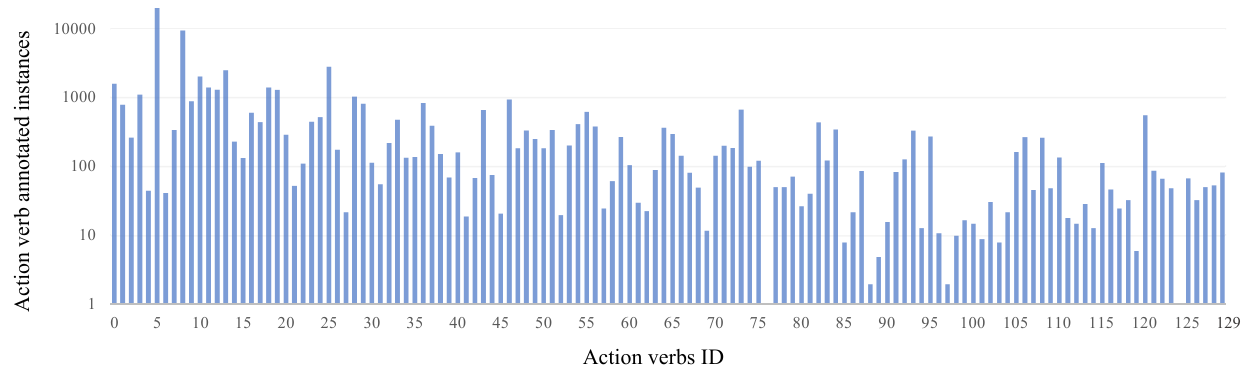}
  \caption{The distribution of instances per action verb category (the outer ring in Figure~\ref{fig:cakefig-actionverbs}) in the FHA-Kitchens dataset.} \label{fig:actionverbdistribution}
  \includegraphics[width=8.3cm,height=2.8cm]{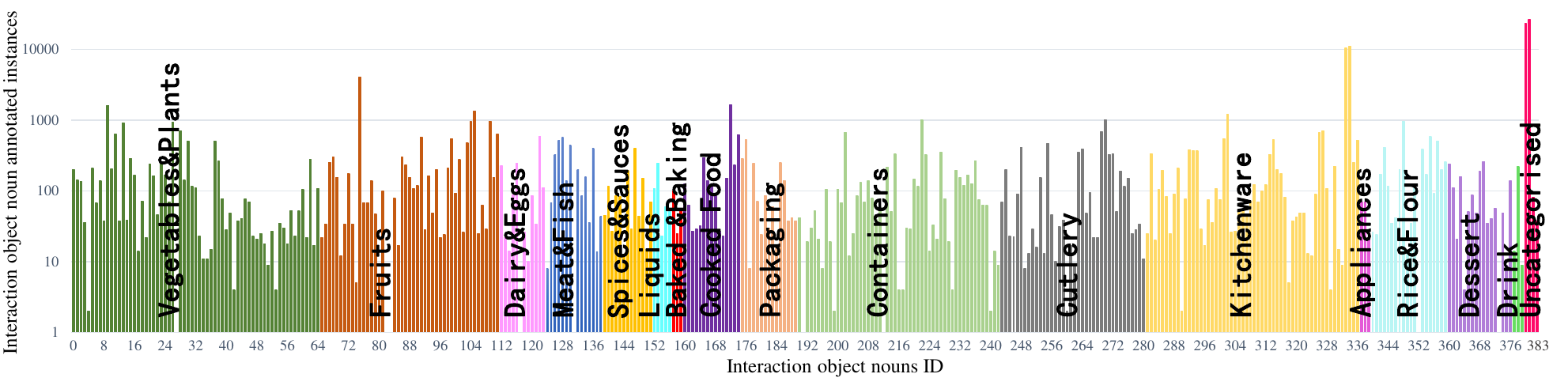}
  \caption{The distribution of instances per object noun category from 17 super-categories in the FHA-Kitchens dataset.} 
  \label{fig:objectnoundistribution}
 \end{minipage}
\end{figure}

\textbf{Annotation Statistics.} 
Our annotation primarily focuses on hand interaction regions, interaction objects, and their corresponding interaction actions, resulting in a diverse array of verbs, nouns, and bounding boxes. Following the fine-grained annotation principles~\cite{1epic}, we ensured minimal semantic overlap among action verb-noun categories, rendering them suitable for multi-category action recognition and detection. \textbf{(1) Verbs:} The annotated dataset comprises 130 action verbs that have been grouped into 43 parent verb categories (Figure~\ref{fig:cakefig-actionverbs} and Figure~\ref{fig:actionverbdistribution}). The three most prevalent parent verb categories, based on the count of sub-action verbs, are \textit{Cut}, \textit{Hold}, and \textit{Take}, representing the most frequently occurring hand actions in kitchen scenes. Figure~\ref{fig:actionverbdistribution} visually depicts the distribution of all verb categories within FHA-Kitchens, ensuring the presence of at least one instance for each verb category. \textbf{(2) Nouns:} In the annotation process, we identified a total of 384 interaction object noun categories that are associated with actions, categorized into 17 super-categories. Figure~\ref{fig:objectnoundistribution} shows the distribution of noun categories based on their affiliations with super-categories. Notably, the super-category ``vegetables \& plants'' exhibits the highest number of sub-categories, followed by ``kitchenware'', which aligns with typical kitchen scenes. \textbf{(3) Bounding Boxes:} We performed a comprehensive statistical analysis on the bounding boxes of the three sub-interaction regions and the corresponding interaction objects. Specifically, we focused on two aspects: the box area and the aspect ratio. Details can be found in \emph{supplementary material}.

\subsection{Benchmark Setup}
\textbf{Methods.} We benchmark several representative action recognition methods~\cite{22slowfast,23videoswin,19TSN,videomaev2,hiera} and detection methods~\cite{24faster,50yolox,defordetr, dino} with different backbone networks on the proposed FHA-Kitchens dataset based on the MMAction2~\cite{51mmaction2} and MMDetection~\cite{52mmdetection} codebases. We establish three tracks using the provided dataset. \textbf{SL-AD Track}: The aim is to evaluate the supervised learning performance of different detection models on hand interaction regions with different granularity levels of action categories. The results of the methods are shown in Table~\ref{tab:SL-AD}. \textbf{SL-AR Track}: This track primarily evaluates the supervised learning performance of different action recognition models on fine-grained hand actions. We trained the models with and without pre-trained weights on the FHA-Kitchens dataset. \textbf{DG Track}: It focuses on experiments for Intra- and Inter-class Domain Generalization in Interaction Region Detection, exploring both intra-class and inter-class perspectives. All models on the SL-AD, SL-AR, and DG tracks were trained and tested using NVIDIA GeForce RTX 3090 GPUs. For the SL-AD and DG tracks, we employ the mean Average Precision (mAP)~\cite{53coco} as the primary evaluation metric, while for the SL-AR track, Top-1 accuracy and Top-5 accuracy (\%) are adopted. Detailed results of SL-AR and DG can be found in \emph{supplementary material}.

\textbf{Results and Discussion.} The results in Table~\ref{tab:SL-AD} show that current detection methods perform well in learning single-dimensional coarse-grained categories like verbs or nouns. However, they struggle in learning multi-dimensional fine-grained action categories. Understanding the intricate nature of real-world hand actions, which encompass both coarse- and fine-grained information, underscores the significance of investigating multi-granularity action categories in HAD tasks, an area that poses significant challenges and remains largely unexplored. To fill this gap, we propose a novel method for multi-granularity hand action detection.

\begin{table}[t!]
  \caption{Detection results (mAP) of hand interaction regions with different granularity levels of action categories using different methods, \textit{i.e.}, Faster-RCNN, YOLOX, Deformable DETR, and DINO on the validation set of the SL-D track.}
  \label{tab:SL-AD}
  \centering
  \renewcommand\arraystretch{1}
  \scalebox{0.8}{
  \begin{tabular}{cccc}
    \toprule
    \multirow{2}{*}{Method}&\multirow{2}{*}{Backbone}&\multicolumn{2}{c}{Granularity levels}\\
     \cmidrule(r){3-4}
     \multirow{2}{*}{~} & \multirow{2}{*}{~} &Coarse-Grained & Fine-Grained \\
     \midrule
     \multirow{2}{*}{Faster-RCNN~\cite{24faster}} & R-50& 65.2 & 48.5\\
     \multirow{2}{*}{~}& R-101 & 66.1 & 50.0 \\
     \multirow{2}{*}{YOLOX~\cite{50yolox}} & YOLOX-s & 71.8 & 46.9 \\
     \multirow{2}{*}{~}& YOLOX-x & 75.6 & 49.8  \\
     Deformable DETR~\cite{defordetr} & R-50 & 73.0 & 52.4 \\
     DINO~\cite{dino} & R-50 & 75.2 & 53.5 \\
  \bottomrule
\end{tabular}}
\end{table}

\begin{figure*}[htbp]
\centering
\includegraphics[width=0.92\linewidth]{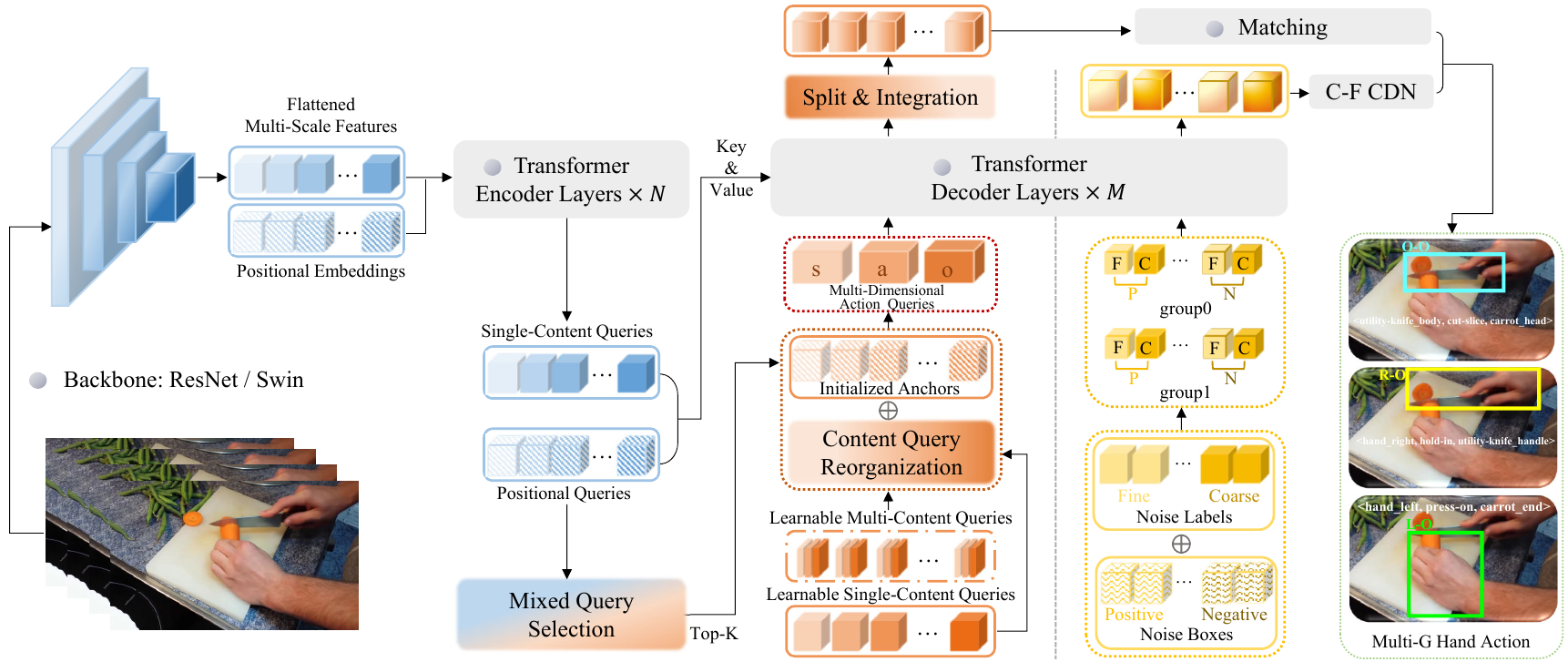}
\caption{The overall architecture of MG-HAD, a novel end-to-end hand action detection model based on DINO~\cite{dino}. The improvements mainly focus on the decoder part. Specifically, (1) we introduce a new design for the content query part, transforming the original single-dimensional content queries into multi-dimensional ones. They are further processed by the designed CQR module, combined with initialized anchors, and inputted into the decoder. The outputted three query sets with different action dimensions go through the Split \& Integration module to generate $N$ queries containing three action dimensions. Finally, the matching process is conducted to predict hand action results (see Section~\ref{4.2}); (2) we introduce a C-F CDN training approach, which involves adding coarse- and fine-grained noise to labels to generate four types of CDN queries for contrastive denoising training (see Section~\ref{4.3}). F: Fine-grained, C: Coarse-grained, Multi-G: Multi-granularity.}
\label{fig:method}
\end{figure*}

\section{A Simple Yet Strong Baseline}
\subsection{A Multi-Granularity Framework} 
Drawing inspiration from the image-based DINO~\cite{dino}, we propose the novel MG-HAD method with specific novel designs in the decoder for multi-granularity hand action detection (Figure~\ref{fig:method}). MG-HAD consists of a backbone, a multi-layer Transformer encoder, a multi-layer Transformer decoder, and multiple prediction branch heads. Given a video clip, for each frame, we utilize backbones like ResNet~\cite{restnet} or Swin Transformer~\cite{swin} to extract multi-scale features, which are then fed into the Transformer encoder along with corresponding positional embeddings. After enhancing features through the encoder layers, we initialize anchors as positional queries for the decoder using a mixed query selection strategy, following the design of DINO, without initializing content queries but leaving them learnable. It's worth noting that the original content queries focus on semantic information of single-dimensional categories, which is not suitable for fine-grained multi-dimensional categories in the new task. Therefore, we modify the single-content query to multi-content queries and introduce a Content Query Reorganization (CQR) module to obtain query sets focusing on three different sub-action dimensions, as detailed in Section~\ref{4.2}. Additionally, similar to DINO, we have an extra CDN branch to perform contrastive denoising training. In contrast to the standard CDN method, we specifically devise a novel coarse-fine granularity contrastive denoising training approach to distinguish labels with different granularity levels, which will be discussed in Section~\ref{4.3}.

\subsection{Multi-Dimensional Action Queries}~\label{4.2}
Comparing the results presented in Table~\ref{tab:SL-AD}, it's clear that existing detection methods struggle with learning from multi-dimensional fine-grained labels. Fine-grained hand action detection poses a greater challenge compared to coarse-grained detection due to the need to discern subtle differences in similar hand actions. Additionally, multi-dimensional fine-grained labels provide important supervisory signals about subject, object, and action categories, as well as localization information. However, effectively encoding this information at different dimensions and leveraging these supervisory signals, particularly in terms of query design within the DETR series framework~\cite{detr,defordetr,dino}, remains unexplored.

\textbf{Implementation:} We observed that the current design of content queries mainly focuses on single-dimensional semantic information, \textit{i.e.}, single verb or noun categories. However, in fine-grained categories, we incorporate both verb and noun categories, generating multi-dimensional semantic information, \textit{i.e.}, \emph{<$c_1$, $c_2$, $c_3$>} ($c_1$, $c_3$ $\in$ \emph{nouns}, and $c_2$ $\in$ \emph{verbs}), or more specific \emph{<s, a, o>}. If we stick to the original design, content queries would consider \emph{<s, a, o>} as a whole, learning global information from a single-dimensional perspective. To enhance the model's focus on local information of sub-categories, we transform a set of content queries $Q = \{q_1, ..., q_n\}$ originally focusing on single dimensions into three sets of content queries, \textit{i.e.}, $Q_s$, $Q_a$, and $Q_o$, focusing on different action dimensions. $n$ is the index of the original queries. Specifically, we first convert each query element $q_n$ (\emph{bottom orange cubes in Figure~\ref{fig:method}}) into three sub-queries, \textit{i.e.}, $q_{n_s}$, $q_{n_a}$, and $q_{n_o}$, expanding $N$ original queries to ${3 \times N}$ sub-queries. Next, through our designed Content Query Reorganization (CQR) module, sub-queries focusing on the same action dimension (\textit{i.e.}, $q_{1_s}$, $q_{2_s}$, ..., $q_{n_s}$) are selected and reorganized to obtain a query set for each action dimension. Additionally, to ensure a comprehensive understanding of fine-grained categories, we introduce an action dimensional hyper-parameter $w_d$ ($d$ $\in$ \{s, a, o\}), to add a certain proportion of weight to each query set, which is then sum with the global information (\textit{i.e.}, $Q = \{q_1, ..., q_n\}$). This process is formulated as:

  \begin{equation}
  Q_d = CQR\left(\left\{q_{n_d}\right\}_{n=1}^{N}\right)=\sum_{n=1}^{N}q_{n_d}\ \ \times\ w_d+Q , \ d = s, a, or \ o.
 \end{equation}

After passing through the CQR module, we obtain content query sets for the three action dimensions. These sets are then summed with the initialized anchors to yield multi-dimensional action queries. Each dimensional query set has a length of $N$, resulting in a total length of $3N$. Following the decoder layers, three query sets for different action dimensions ($Q_s'$, $Q_a'$, and $Q_o'$) are outputted. Subsequently, through the Split\&Integration module, queries from different action dimensions with the same index (\textit{e.g.}, $q_{1_s}$, $q_{1_a}$,$q_{1_o}$) are integrated to generate $N$ queries, each of which contains information from three action dimensions (\emph{top orange cubes in Figure~\ref{fig:method}}): 
\begin{equation}
  \left\{q_n\right\}_{n=1}^N=SI\left(Q_s', Q_a', Q_o'\right)=\ \left\{q_{n_s}+q_{n_a}+q_{n_o}\right\}_{n=1}^N ,
 \end{equation}
where \emph{$SI$} represents the Split\&Integration module. Finally, the matching process is conducted to predict hand action results. 

\textbf{Analysis:} In our design of multi-dimensional action queries, we introduce an action dimensional hyper-parameter $w_d$ ($d$ $\in$ \{s, a, o\}), to control the proportion of local information (sub-categories) fused with global information (triplet categories). In the three action dimensions \emph{<s, a, o>}, the \emph{a} dimension is the most crucial for our task. Therefore, we use $w_a$ as the central weight to dynamically adjust the weight proportions of the three action dimensions, with a total sum of 1. To determine the optimal weight distribution, we conducted a total of 10 comparative experiments under different backbones, with detailed results provided in \emph{supplement material}.

\subsection{Coarse-Fine Contrastive Denoising}~\label{4.3}
For object detection, DINO is highly effective in stabilizing training and accelerating convergence. With the help of DN queries, it learns to predict ``no object'' for anchors without nearby objects, thereby inhibiting confusion and selecting high-quality anchors (queries) for predicting bounding boxes. However, in HAD tasks where hand action categories may overlap or be similar, DINO primarily addresses the confusion of boxes but overlooks label categories, resulting in poor prediction capability for different granularity levels of hand action categories. To address this issue, we propose a \textbf{C}oarse-\textbf{F}ine granularity \textbf{C}ontrastive \textbf{D}e\textbf{N}oising (C-F CDN) training approach to reject anchors with ``incorrect granularity labels''.

\textbf{Implementation:} DINO introduces two hyper-parameters $\gamma_1$ and $\gamma_2$, to control the scale of box and label noise, respectively. The generated noises are no larger than $\gamma_1$ and $\gamma_2$, aiming to enable the model to reconstruct the ground truth (GT) from moderately noisy queries. We observed that DINO only designs two types of CDN queries for the box: positive and negative queries, while the label is set to be randomly generated. In the proposed method, while keeping the box settings unchanged, we further generate two types of CDN queries for the label: coarse-grained and fine-grained queries (\emph{dark and light yellow cubes in Figure~\ref{fig:method}}). Moreover, unlike the strategy of randomly generating noisy labels, we add noise by specifying the noise position and noise category for different granularity labels. Specifically, coarse-grained queries add noise containing fine-grained information, while fine-grained queries add noise containing coarse-grained information, with the expectation of predicting the correct granularity label for each GT box. In Figure~\ref{fig:method}, each CDN group comprises four types of queries: positive-coarse, positive-fine, negative-coarse, and negative-fine. If a frame has $n$ GT bounding boxes, a CDN group will contain 4 types of ${2 \times n}$ queries. Similar to DINO, we also utilize multiple CDN groups to enhance the effectiveness of the method. The reconstruction loss for bounding box regression includes $\mathit{l}_1$ and GIOU losses, while focal loss~\cite{focalloss} is employed for classification.

\textbf{Analysis:} When designing the noise label generation strategy, we replaced the ``random'' generation of noise with ``specified'', reducing randomness by specifying noise positions and categories. This ensures noise is added to different granularity labels, generating CDN queries encompassing various granularity. To determine the optimal setting, we considered the noise distribution of different granularity categories in real-world scenarios and ensured contrastive learning between coarse and fine-grained information. We conducted three sets of comparative experiments (Details can be found in \emph{supplementary material}). The final selected setting, as shown in Eq.~(\ref{equ:c-f}), exhibits the most significant improvement. Hence, subsequent experiments were conducted using this setting for further investigation. Our method's success lies in its ability to suppress confusion at the category level and select appropriate granularity to predict hand action categories, thus enhancing its ability to predict multi-granularity information.
    \begin{equation}\label{equ:c-f}
    \text{noise label} = \left\{
    \begin{array}{ll}
    \text{fine-grained} & i \in [0,3) \\
    \text{mixed-grained} & i \in [3,C)
    \end{array}
    \right.,
    \end{equation}
where $i$ indexes the multi-granularity action category for a specific instance (\textit{i.e.}, 0$\sim$2 denote coarse-grained categories while 3$\sim$C-1 denote fine-grained categories), and $C$ is the number of categories. ``fine-grained'' and ``mixed-grained'' denote that the noise label is chosen randomly from the fine-grained categories and the combination of the coarse-grained and fine-grained categories, respectively.

\begin{table}
  \caption{Results for MG-HAD and other DETR Series detection models with the ResNet50 backbone on the FHA-Kitchens validation set trained with \emph{12} epochs. M-G: Mixed-Grained, C-G: Coarse-Grained, F-G: Fine-Grained.}
  \label{tab:12epoch}
  \centering
  \scalebox{0.85}{
  \begin{tabular}{cccc}
    \toprule
     \multirow{2}{*}{Method} & \multicolumn{3}{c}{FHA-Kitchens val mAP(\%)}\\
     \cmidrule(r){2-4}
     \multirow{2}{*}{~}& M-G label& C-G sub-label& F-G sub-label \\
     \midrule
     \emph{DETR}~\cite{detr} & 42.3 & 72.8 & 41.9\\
     \emph{Deformable DETR}~\cite{defordetr}&  49.4 & 70.9 & 49.1 \\
     \emph{DAB-DETR}~\cite{dab} & 52.1 & 73.1 & 51.8\\
     \emph{DDQ-4scale}~\cite{ddq}& 53.8 & 67.8 & 53.7 \\
     \emph{DINO-4scale}~\cite{dino} & 54.7 & 76.3 & 54.5 \\
     \textbf{\emph{MG-HAD-4sacle}} & \textbf{57.0}($+$2.3) & \textbf{75.6} & \textbf{56.8}($+$2.3)  \\
  \bottomrule
\end{tabular}}
\end{table}

\section{Experiments}
\subsection{Experiments Settings}
\textbf{Dataset and Metric.} Due to the absence of benchmarks for this new task, we evaluated our model and other representative detection models solely on the FHA-Kitchens dataset. We conducted experiments using two different backbones: ResNet-50~\cite{restnet} pre-trained on ImageNet-1k~\cite{imagenet} and Swin-L~\cite{swin} pre-trained on ImageNet-22k~\cite{imagenet}. All detection models utilize pre-trained weights on the MS COCO object detection dataset~\cite{53coco}. Furthermore, we not only report the overall validation results using mixed-grained labels but also separately report the validation results for coarse-grained and fine-grained sub-labels. We follow previous works and adopt mean Average Precision (mAP)~\cite{53coco} as the primary evaluation metric.
  
\textbf{Implementation Details.} We trained the MG-HAD model on the FHA-Kitchen dataset using the MMDetection~\cite{52mmdetection} codebase. Specifically, we utilized pre-trained weights on the MS COCO~\cite{53coco} object detection dataset and fine-tuned it on the hand action detection task on FHA-Kitchens. We trained the model under two different settings: 4scale-R-50 and 5scale-Swin-L. Following DINO~\cite{dino}, we used the Adam optimizer~\cite{adam} for model training, with an initial learning rate of ${1 \times 10^{-4}}$ and weight decay is $10^{-4}$. The experiments were conducted on the NVIDIA GeForce RTX 3090 GPUs, with a batch size of 2 for 4scale-R-50 and 1 for 5scale-Swin-L. By default, MG-HAD was trained for 12 epochs, taking approximately 5 hours. More details are provided in the \emph{supplement material}.

\begin{table}
 \caption{Results of MG-HAD and other SOTA detection models on the FHA-Kitchens validation set. R-50: ResNet-50, M-G: Mixed-Grained, C-G: Coarse-Grained, F-G: Fine-Grained.}
    \centering
    \scalebox{0.72}{
    \begin{tabular}{cccccc}
    \toprule
         \multirow{2}{*}{Method}& \multirow{2}{*}{Epoch}&  \multirow{2}{*}{Backbone}& \multicolumn{3}{c}{FHA-Kitchens val mAP(\%)}\\
    \cmidrule(r){4-6}   
         \multirow{2}{*}{~}& \multirow{2}{*}{~}&  \multirow{2}{*}{~}& M-G label & C-G sub-label & F-G sub-label\\
    \midrule
         \emph{Faster R-CNN}~\cite{24faster}& 108 & R-50 & 48.3	& 22.3 & 48.6 \\
         \emph{YOLOX}~\cite{50yolox} & 100 & YOLOX-x & 50.7 & 70.8 & 50.5 \\     
         \emph{DETR}~\cite{detr}& 150 & R-50& 50.6 & 73.1 & 50.3 \\
         \emph{Deformable DETR}~\cite{defordetr}& 50 & R-50 & 53.7 & 72.6 & 53.4 \\
         \emph{DAB-DETR}~\cite{dab}& 50 & R-50 & 54.7 &	75.2 & 54.5 \\
         \emph{DINO-4scale}~\cite{dino}& 24 & R-50 & 56.3 & 74.5 & 56.0 \\
         \emph{DINO-4scale}~\cite{dino}& 12 & R-50 & 54.7 & 76.3 & 54.5 \\
         \emph{DINO-5scale}~\cite{dino}& 12 & Swin-L & 56.3 & 76.3 & 56.1\\ 
    \cmidrule(r){1-6} 
        \multirow{2}{*}{\textbf{\emph{MG-HAD-4scale}}} & 24 & \multirow{2}{*}{R-50} & \textbf{57.7}($+$1.4) & \textbf{75.3} & \textbf{57.5}($+$1.5) \\
         \multirow{2}{*}{~} & 12 & \multirow{2}{*}{~} & \textbf{57.0}($+$2.3) & \textbf{75.6} & \textbf{56.8}($+$2.3)\\
     \cmidrule(r){1-6} 
         \textbf{\emph{MG-HAD-5scale}} & 12 & Swin-L & \textbf{59.4}($+$3.1) & \textbf{77.6}  & \textbf{59.2}($+$3.1) \\
    \bottomrule
    \end{tabular}}
    \label{tab:sotaresult}
\end{table}
  
\subsection{Main Results}
\textbf{12-Epoch Setting.} To demonstrate the effectiveness of our method for the multi-granularity HAD task, we compared it with representative strong baselines from the DETR series~\cite{detr,defordetr,dab,ddq,dino} on the FHA-Kitchens dataset under the setting of ResNet-50 backbone and 12 epochs. In particular, our method, DINO~\cite{dino}, and DDQ~\cite{ddq} mainly report results under the 4scale setting. As shown in Table~\ref{tab:12epoch}, our method achieves much better accuracy in detecting mixed-grained hand actions, owing to the proposed C-F CDN module and multi-dimensional action queries. Specifically, it achieves an improvement of \emph{\textbf{$+$2.3}} AP on mixed-grained labels compared to the current strongest baseline DINO~\cite{dino} under the same setting. Furthermore, compared to the classic DETR~\cite{detr}, our method achieves a significant improvement of \emph{\textbf{$+$14.7}} AP. Note that our method not only performs well for mixed-grained labels but also shows improvement in the validation results for fine-grained sub-labels.

\textbf{Comparison with SOTA Detection Methods.} 
To comprehensively and fairly validate the effectiveness of our method in enhancing the performance of multi-granularity hand actions, we compared it with other state-of-the-art (SOTA) detection methods on the FHA-Kitchens dataset, utilizing their optimal settings (refer to the MMDetection~\cite{52mmdetection} codebase). DINO exhibits relatively fast convergence, achieving good results with just 12 epochs on the Swin-L backbone. Our method inherits the convergence capability of DINO but yields more significant improvements. We adopted the same settings as DINO~\cite{dino}, utilizing both 4scale ResNet-50 and 5scale Swin-L backbones, trained for 12 epochs and 24 (2$\times$) epochs, respectively. The results in Table~\ref{tab:sotaresult} indicate the following: \textbf{(1)} Our method exhibits a significant improvement compared to the baseline~\cite{dino}, which can be attributed to the design of handling fine-grained information in our model; \textbf{(2)} Comparing models trained for 24 epochs and 12 epochs, the main improvement lies in the accuracy of fine-grained action detection. Since the FHA-Kitchens dataset contains overwhelming fine-grained categories over the coarse-grained ones, the model's representation capacity may be primarily utilized for fitting fine-grained categories; \textbf{(3)} Under the 5-scale Swin-L backbone, our method achieves a significant improvement of \emph{\textbf{59.4}} AP for mixed-grained hand actions with just 12 epochs. This indicates that using a more powerful backbone~\cite{swin} can improve both coarse- and fine-grained action detection accuracy. Specifically, the detection accuracy of fine-grained actions is increased by \emph{\textbf{$+$3.1}} AP and the detection accuracy of coarse-grained actions is increased by \emph{\textbf{$+$1.3}} AP. The visualization of the detection results can be found in the \emph{supplement material}.

\begin{table}
  \caption{Ablation study of the key components in MG-HAD. C-F CDN: Coarse-Fine granularity Contrastive De-Noising Training, Multi-DA Q: Multi-Dimensional Action Queries. }
  \renewcommand\arraystretch{1.3}
      \centering
      \scalebox{0.8}{
      \begin{tabular}{c|cc|cc}
      \toprule
           \textbf{Method} &\multicolumn{2}{c}{Algorithm Components} & \multicolumn{2}{c}{mAP(\%)}\\
       \midrule
           & \emph{C-F CDN} & \emph{Multi-DA Q} & \emph{4scale-R-50} & \emph{5scale-Swin-L}\\
           \cline{2-5} 
           \textbf{Baseline}~\cite{dino} &&& 54.7 & 56.3  
           \\
           \cline{1-5} 
           \multirow{3}{*}{~}& \textbf{\checkmark} && 56.6 & 58.2  \\
           \multirow{3}{*}{\textbf{MG-HAD}}&& \textbf{\checkmark} & 56.4 & 58.7  \\
           \multirow{3}{*}{~}& \textbf{\checkmark} &\textbf{\checkmark}& \textbf{57.0} & \textbf{59.4}  \\
        \bottomrule
      \end{tabular}}
      \label{tab:ablation}
  \end{table}

\subsection{Ablation Studies} 
\textbf{Effectiveness of New Components:}
Our method utilizes the multi-dimensional action queries for multi-dimensional information processing, as introduced in Section~\ref{4.2}, and the C-F CDN module for multi-granularity information processing, as described in Section~\ref{4.3}. To further validate the effectiveness of these components, we separately isolated them from the model and evaluated the performance under two settings: 4scale ResNet-50 and 5scale Swin-L, as shown in Table~\ref{tab:ablation}, where the baseline denotes the original design proposed by DINO~\cite{dino}. As can be seen, while the strong baseline DINO~\cite{dino} has already surpassed previous models, the proposed MG-HAD introduces two novel designs that notably boost performance in hand action detection. Each module significantly enhances the baseline on both backbones, and their combined effect further enhances performance, demonstrating their complementary role in understanding multi-granularity hand action information.

\section{Conclusion}
In this paper, we present the first study on multi-granularity hand action detection, aiming to understand the diverse hand actions through localizing regions and recognizing various granularity categories of hand actions. We establish \textbf{FHA-Kitchens}, the first fine-grained hand action detection dataset, comprising 30,047 high-quality video frames, 198,839 bounding boxes, and 880 hand action categories. Through systematic evaluation, we identify that existing detection methods excel in coarse-grained actions but struggle with fine-grained ones. To address this, we propose \textbf{MG-HAD}, a simple yet strong baseline model leveraging the Transformer detector with two novel designs. It outperforms previous methods across various granularities of hand actions. FHA-Kitchens and MG-HAD can serve as a valuable testbed and baseline for future research.

\section*{ACKNOWLEDGMENTS}
This work was supported in part by the National Key Research and Development Program of China under No. 2021YFC3300200, the National Natural Science Foundation of China (Grant No. 62276195), and the Special Fund of Hubei Luojia Laboratory under Grant 220100014. The numerical calculations in this paper were done using the supercomputing system at the Supercomputing Center of Wuhan University. We also thank all the volunteers who contributed to the dataset.

\balance

\newpage
\appendix

\section{Appendix}

\subsection{FHA-Kitchens Dataset}

\subsubsection{Data Annotation}
\

 We recruited 10 voluntary annotators to annotate hand actions for each frame with high quality. Their responsibility was to annotate bounding boxes and multi-granularity action categories for each hand interaction region. To enhance annotation efficiency, we implemented a parallel annotation approach. We utilized the \href{https://github.com/open-mmlab/labelbee-client}{LabelBee} tool for annotating bounding boxes and coarse-grained categories, while fine-grained action triplets were annotated on the \href{https://www.mturk.com/}{Amazon Mechanical Turk} platform. To ensure annotation quality, we conducted three rounds of cross-checking and corrections. In the main paper, we described the annotation details for bounding boxes and categories. In addition to this, we also provide segment annotations for the objects.

\begin{figure}[b]
  \centering
  \includegraphics[width=7.5cm]{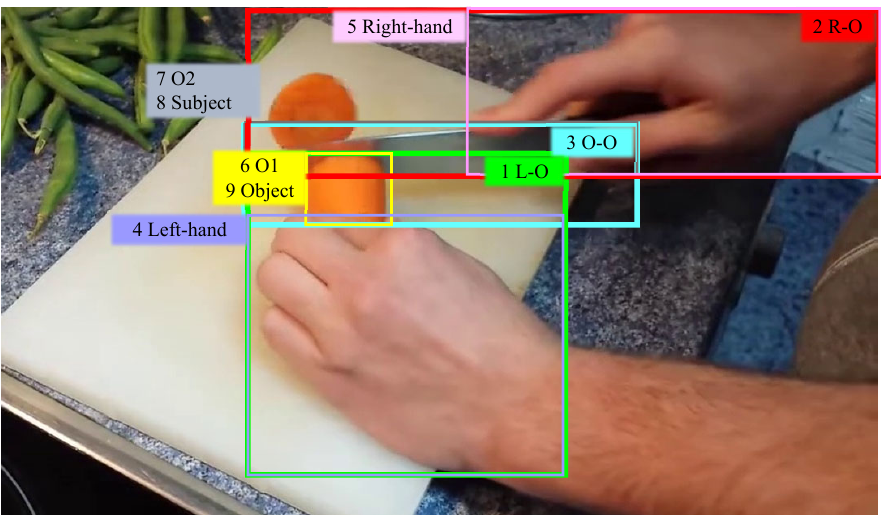}
    \caption{Visualization of bounding box annotations for the example of ``\emph{fry vegetables}''.}
    \label{annvis}
  \end{figure}

\begin{figure}[b]
  \centering
  \includegraphics[width=7.5cm]{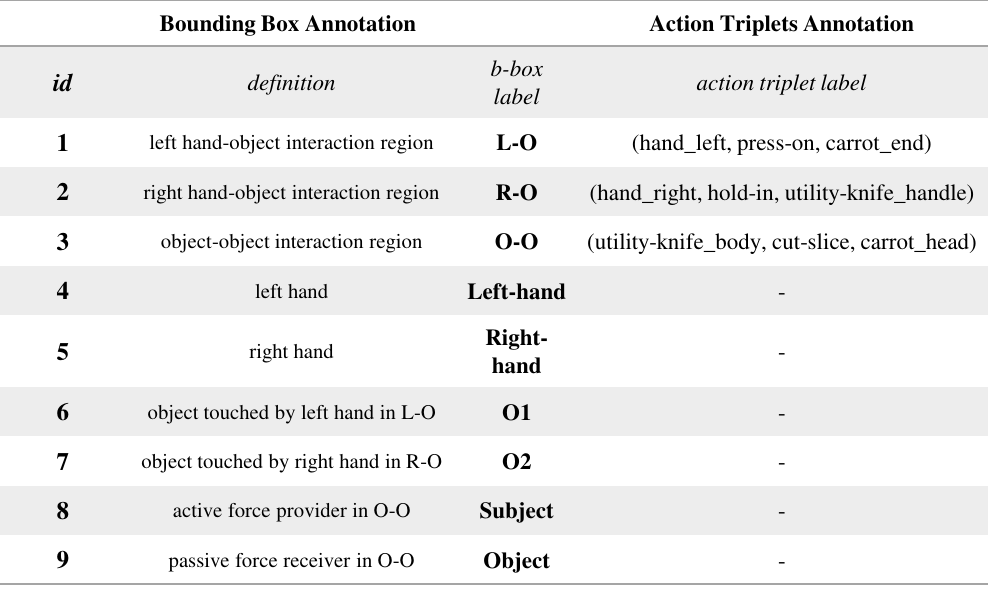}
  \caption{Descriptive list of action triplets and bounding box annotations.}
  \label{annlist}
  \end{figure}

  \textbf{Bounding Box Annotation.} During annotation, we may encounter overlapping bounding boxes, \textit{i.e.}, the same interacting object will satisfy two annotation definitions, for example, the \textit{utility knife} in Figure~\ref{annvis}, which is both the object directly touched by the right hand in the R-O interaction region and the active force provider in the O-O interaction region. In this case, we annotated all the labels because the same object participates in different interaction actions and has different roles (The corresponding visualizations are shown in Figure~\ref{annvis} and the annotation details are listed in Figure~\ref{annlist}). Finally, we annotated a total of 198,839 bounding boxes over 9 types, including 49,746 hand boxes, 66,402 interaction region boxes, and 82,691 interaction object boxes. Compared to existing datasets~\cite{1epic}, we added an average of 5 additional annotation types per frame.

 \textbf{Object Segment Annotation.} To enrich our FHA-Kitchens, we utilized the state-of-the-art SAM model~\cite{20segment} to annotate object masks in all video frames, which can be used for action segmentation relevant tasks.

\subsubsection{More statistics of the FHA-Kitchens Dataset}~\label{statistics}
\

 In this part, we re-arrange some figures in the paper to make them more readable and provide more statistics of the FHA-Kitchens dataset. Our annotation primarily focuses on hand interaction regions, interaction objects, and their corresponding interaction actions, resulting in a diverse array of verbs, nouns, and bounding boxes.

\begin{figure*}[ht]
  \centering
  \includegraphics[width=0.52\linewidth]{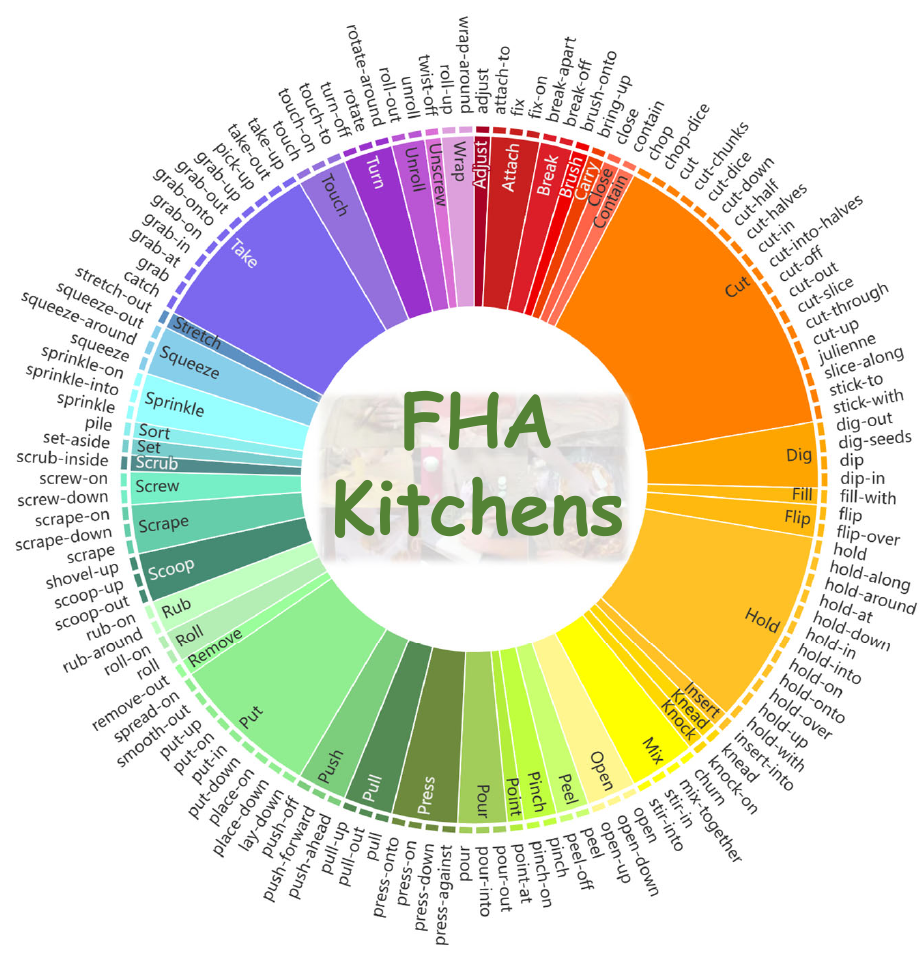}
  \caption{An overview of the action verbs and their parent action categories in FHA-Kitchens.}
  \label{sup_cakefig-actionverbs}
\end{figure*}

\begin{figure*}[htbp]
 \centering
  \includegraphics[width=0.9\linewidth]{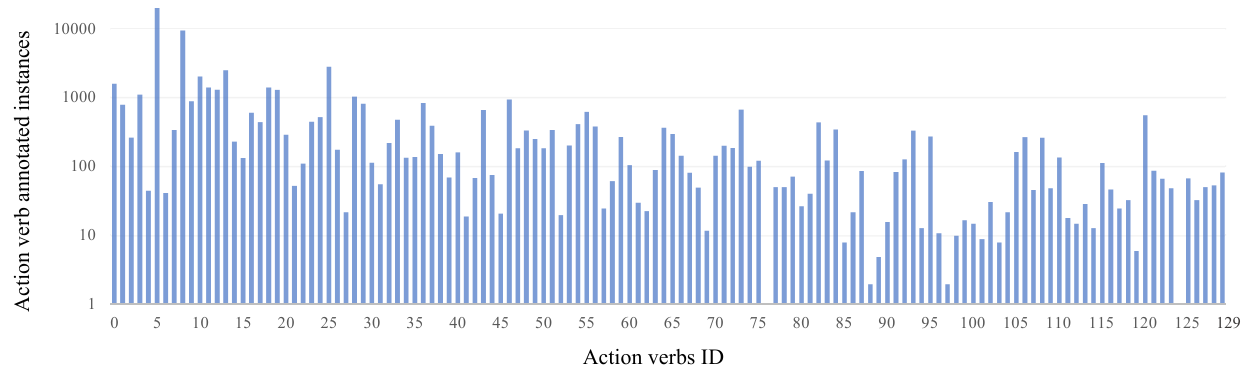}
  \caption{The distribution of instances per action verb category (the outer ring of the circle in Figure~\ref{sup_cakefig-actionverbs}) in the FHA-Kitchens dataset.} \label{sup_actionverbdistribution}
\end{figure*}

 \textbf{Verbs.} The annotated dataset comprises 130 action verbs that have been grouped into 43 parent verb categories (Figure~\ref{sup_cakefig-actionverbs} and Figure~\ref{sup_actionverbdistribution}). The three most prevalent parent verb categories, based on the count of sub-action verbs, are \textit{Cut}, \textit{Hold}, and \textit{Take}, representing the most frequently occurring hand actions in human interactions. Figure~\ref{sup_actionverbdistribution} visually depicts the distribution of all verb categories within FHA-Kitchens, ensuring the presence of at least one instance for each verb category. Specifically, the mapping between action verb IDs and their corresponding category names can be seen in Table~\ref{verbid}. 
 
\begin{figure*}[ht]
  \centering
  \rotatebox{270}{\includegraphics[width=1.3\linewidth]{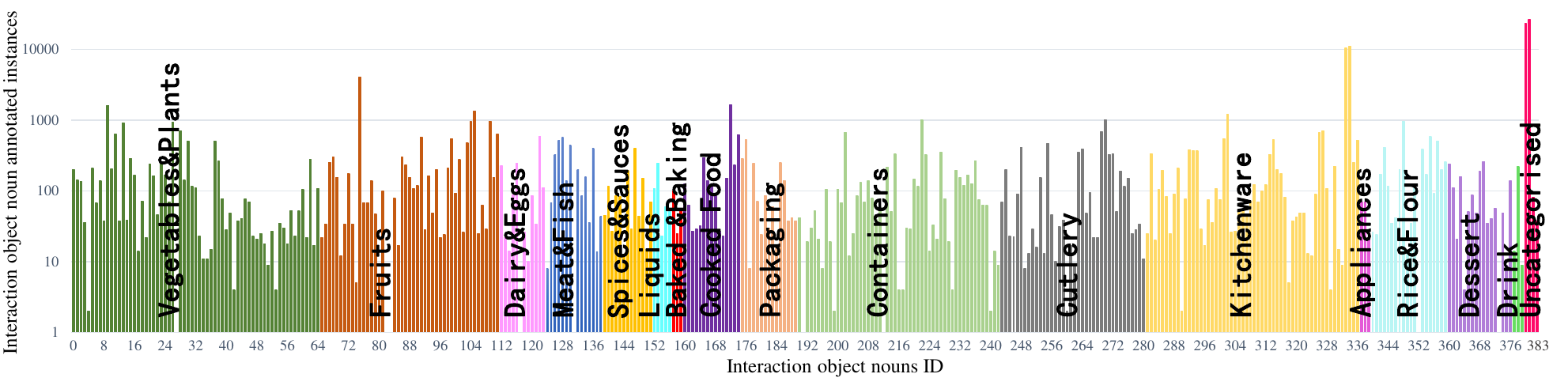}}
  \caption{The distribution of instances per object noun category from 17 super-categories in the FHA-Kitchens dataset.} 
  \label{sup_objectnoundistribution}
\end{figure*}

 \textbf{Nouns.} In our annotation process, we identified a total of 384 interaction object noun categories that are associated with actions, categorized into 17 super-categories. Figure~\ref{sup_objectnoundistribution} shows the distribution of noun categories based on their affiliations with super-categories. Notably, the super-category ``vegetables \& plants'' exhibits the highest number of sub-categories, followed by ``kitchenware'', which aligns with typical kitchen scenes. Specifically, the mapping between interaction object noun IDs and their corresponding category names can be seen in Table~\ref{nounid1}, Table~\ref{nounid2}, and Table~\ref{nounid3}.

 \begin{figure*}[t!]
    \centering
    \includegraphics[width=13.5cm]{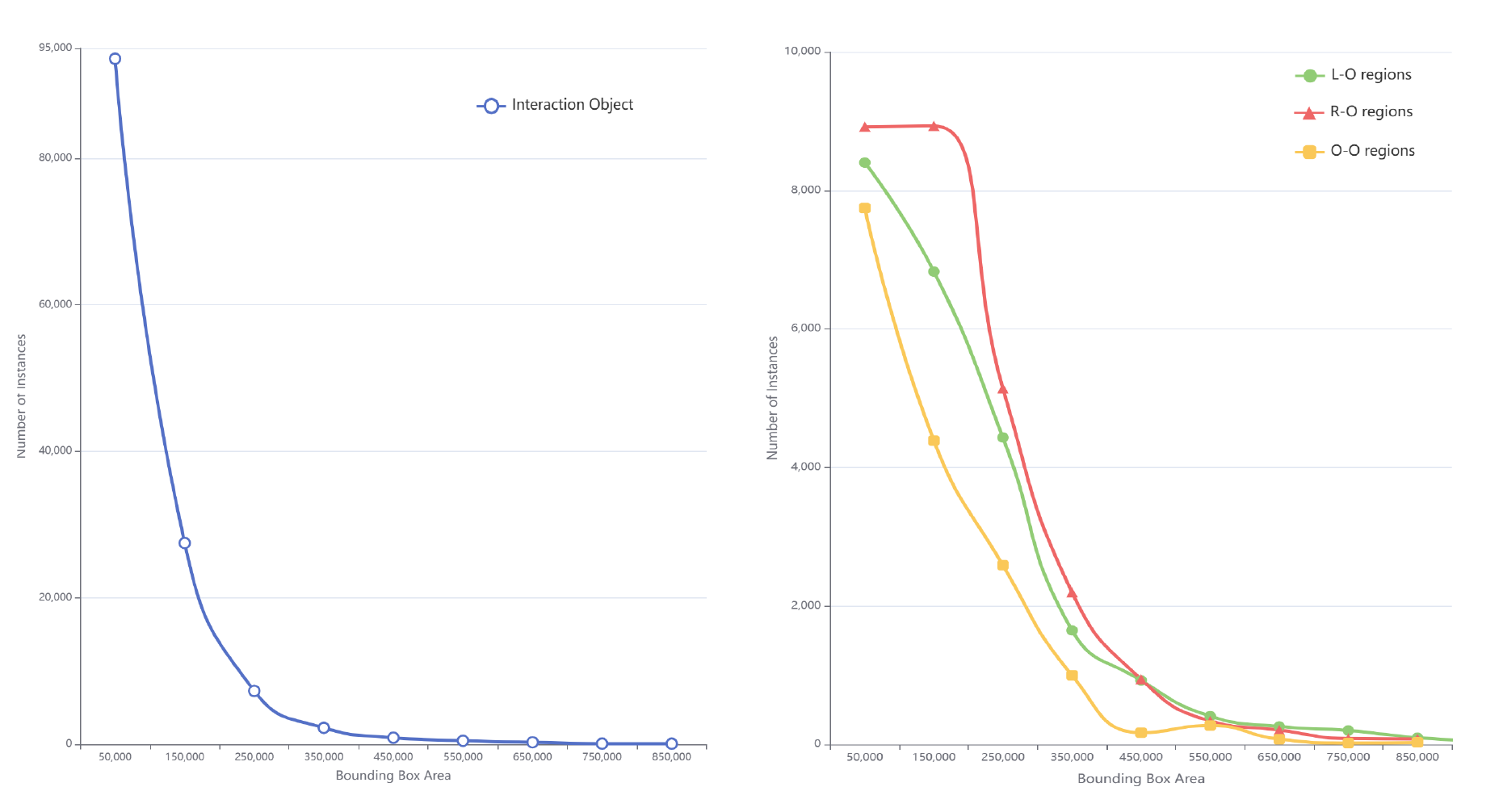}
    \caption{The distributions of bounding box areas of interaction objects (left) and interaction regions (right) in the FHA-Kitchens dataset.}
    \label{area}
\end{figure*}

\begin{figure*}[t!]
  \centering
  \includegraphics[width=13.5cm]{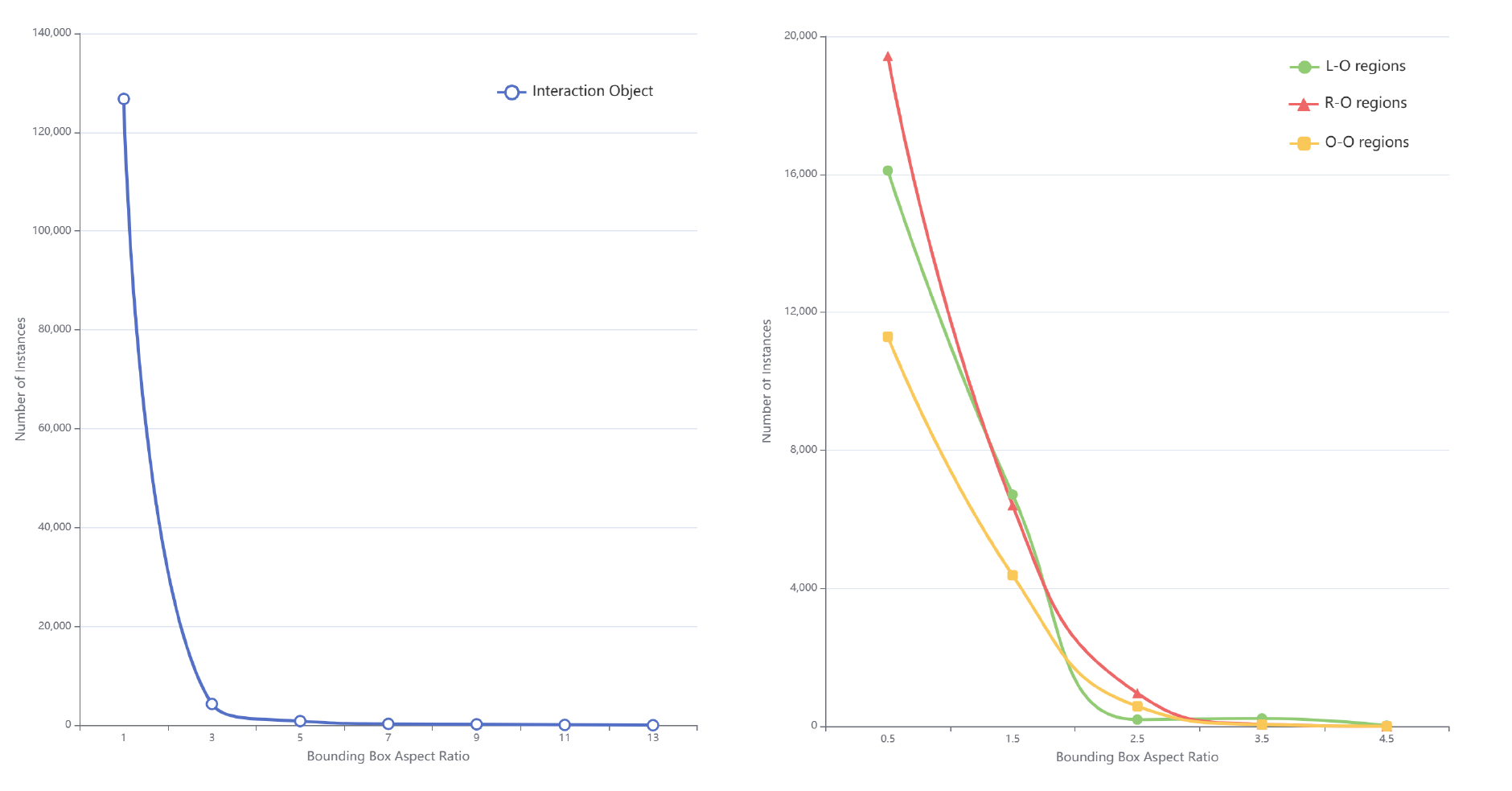}
  \caption{The distributions of bounding box aspect ratios of interaction objects (left) and interaction regions (right) in the FHA-Kitchens dataset.}
  \label{ratio}
\end{figure*}
 
 \textbf{Bounding Boxes.} We performed a comprehensive statistical analysis on the bounding boxes of the three hand interaction regions and the corresponding interacting objects. Specifically, we focused on two aspects: the box area and the aspect ratio. Detailed results can be found in Figure~\ref{area} and Figure~\ref{ratio}. Figure~\ref{area} shows the considerable range of sizes covered by our bounding boxes, with many interaction objects exhibiting small and challenging sizes for accurate detection. Moreover, in Figure~\ref{ratio}, the aspect ratios of the bounding boxes exhibit notable variation. The aspect ratios of the three regions tend to concentrate within the range of [0.5,2], which can be attributed to the typical composition of interaction regions involving two interacting objects. Consequently, the bounding box encompasses the combined region of both objects. For instance, the R-O interaction region frequently involves the interaction between the ``\emph{right hand}'' and ``\emph{utility knife}''. In such cases, the aspect ratio of the bounding box is observed to be 2:1, as depicted in Figure~\ref{annvis}. These findings highlight the significant challenges of the detection task in our dataset.

\begin{figure*}[ht]
  \centering
  \includegraphics[width=0.9\linewidth]{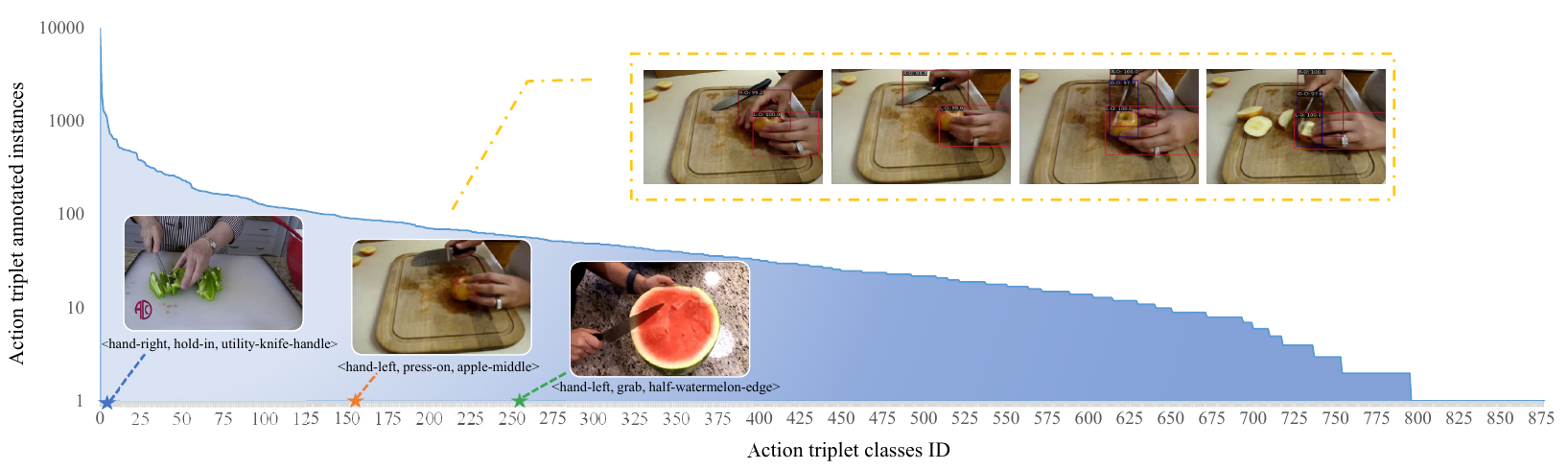}
  \caption{The distribution of instances per action triplet category in the FHA-Kitchens dataset.} 
  \label{actiontripletdistribution}
\end{figure*}

\textbf{Long-tail Property.} 
The distribution of instances per action triplet category in FHA-Kitchens, as depicted in Figure~\ref{actiontripletdistribution}, depicts a long-tail property. This distribution reflects the frequency of hand interactions in real-world kitchen scenes, taking into account the varying commonness or rarity of specific hand actions. For instance, the action triplet ``\emph{<hand\_right, hold-in, utility-knife\_handle>}'' consists of 9,887 instances, which is nine times more prevalent than the ``\emph{<hand\_left, hold-in, utility-knife\_handle>}'' triplet. This long-tail characteristic of the distribution renders FHA-Kitchens a challenging benchmark for hand action recognition, making it suitable for investigating few-shot learning and out-of-distribution generalization in action recognition as well.

\begin{table*}[ht!]
  \centering
  \caption{Classification results (Top-1 and Top-5 accuracy) of fine-grained hand actions using different methods on the validation set of the SL-AR track. w/ Pre-train : using pre-trained weights. w/o Pre-train: Training from scratch (the Kinetics 400~\cite{6kinetic400} dataset results from mmaction2~\cite{51mmaction2}, VideoMAE V2~\cite{videomaev2}, and Hiera~\cite{hiera}).}
  \scalebox{0.86}{
  \begin{tabular}{cccccccc}
  \toprule
  \multirow{2}{*}{Dataset} & \multirow{2}{*}{Method} & \multirow{2}{*}{Backbone} & \multirow{2}{*}{Pre-train Data} & \multicolumn{2}{c}{w/ Pre-train} & \multicolumn{2}{c}{w/o Pre-train}\\
  \cmidrule(r){5-8}
  \multirow{2}{*}{~} & \multirow{2}{*}{~} & \multirow{2}{*}{~} & \multirow{2}{*}{~} & Top-1 & Top-5 & Top-1 & Top-5 \\
  \midrule
  \multirow{9}{*}{Kinetics 400} & \multirow{2}{*} {TSN~\cite{19TSN}} & ResNet50 & ImageNet & 72.83 & 90.65 & - & - \\
  \multirow{9}{*}{~} & \multirow{2}{*}{~} & ResNet101 & ImageNet & 75.89 & 92.07 & - & - \\
  \cmidrule(r){2-8}
  \multirow{9}{*}{~} & \multirow{2}{*}{SlowFast~\cite{22slowfast}} & ResNet50 & - & - & - & 76.65 & 92.86 \\
  \multirow{9}{*}{~} & \multirow{2}{*}{~} & ResNet101 & - & - & - & 78.65 & 93.88\\
  \cmidrule(r){2-8}
 \multirow{9}{*}{~} & VideoSwin~\cite{23videoswin} & Swin-B & ImageNet & 80.57 & 94.49 & - & - \\
 \cmidrule(r){2-8}
  \multirow{9}{*}{~} & \multirow{3}{*}{VideoMAE V2~\cite{videomaev2}} & ViT-B & UnlabeledHybrid & 81.50 & - & - & - \\
  \multirow{9}{*}{~} & \multirow{3}{*}{~} & ViT-L & UnlabeledHybrid & 85.40 & - & - & - \\
  \multirow{9}{*}{~} & \multirow{3}{*}{~} & ViT-H & UnlabeledHybrid & 86.90 & - & - & - \\
 \cmidrule(r){2-8}
 \multirow{9}{*}{~} & Hiera~\cite{hiera} & Hiera-B & Kinetics 400 & 84.00 & - & - & - \\
  \bottomrule
  \toprule
  \multirow{2}{*}{Dataset} & \multirow{2}{*}{Method} & \multirow{2}{*}{Backbone} & \multirow{2}{*}{Pre-train Data} & \multicolumn{2}{c}{w/ Pre-train} & \multicolumn{2}{c}{w/o Pre-train}\\
  \cmidrule(r){5-8}
  \multirow{2}{*}{~} & \multirow{2}{*}{~} & \multirow{2}{*}{~} & \multirow{2}{*}{~}& Top-1 & Top-5 & Top-1 & Top-5 \\
  \midrule
  \multirow{9}{*}{FHA-Kitchens} & \multirow{2}{*}{TSN~\cite{19TSN}} & ResNet50 & Kinetics 400 & 30.37 & 74.26 & 29.11 & 73.84 \\
  \multirow{9}{*}{~} & \multirow{2}{*}{~} & ResNet101 & Kinetics 400 & 30.80 & 73.42 & 30.38 & 74.26 \\
  \cmidrule(r){2-8}
  \multirow{9}{*}{~} & \multirow{2}{*}{SlowFast~\cite{22slowfast}} & ResNet50 & Kinetics 400 & 33.33 & 70.46 & 27.85 & 68.35 \\
 \multirow{9}{*}{~} &  \multirow{2}{*}{~} & ResNet101 & Kinetics 400 & 36.71 & 67.93 & 31.22 & 69.62\\
  \cmidrule(r){2-8}
 \multirow{9}{*}{~} & VideoSwin~\cite{23videoswin} & Swin-B & Kinetics 400 & 37.13 & 70.89 & 34.18 & 66.67 \\
 \cmidrule(r){2-8}
 \multirow{9}{*}{~} & \multirow{3}{*}{VideoMAE V2~\cite{videomaev2}} & ViT-B & UnlabeledHybrid & 21.67 & 57.08 & - & - \\
 \multirow{9}{*}{~} &  \multirow{3}{*}{~} & ViT-L & UnlabeledHybrid & 32.92 & 68.75 & - & - \\
  \multirow{9}{*}{~} & \multirow{3}{*}{~} & ViT-H & UnlabeledHybrid & 34.58 & 68.33 & - & - \\
 \cmidrule(r){2-8}
\multirow{9}{*}{~} & Hiera~\cite{hiera} & Hiera-B & Kinetics 400 & 27.00 & 69.20 & - & - \\
  \bottomrule
 \end{tabular}}
  \label{arfullresults}%
\end{table*}%

\subsubsection{More quantitative and qualitative results}~\label{A.1.3}
\ 

\textbf{1) Quantitative Results}
\

\textbf{SL-AR Track: Supervised Learning for Fine-grained Hand Action Recognition} 

 \textbf{Settings.} The SL-AR track primarily evaluates the performance of different action recognition models on fine-grained hand actions. We adopted the representative TSN~\cite{19TSN} and Slowfast~\cite{22slowfast} with the ResNet50 and ResNet101 backbones, VideoSwin~\cite{23videoswin} with the Swin-B backbone, VideoMAE V2~\cite{videomaev2} with the three different size backbones, and Hiera~\cite{hiera} with the Hiera-B backbone. We trained these models on the FHA-Kitchens dataset using two settings: (1) \textbf{Pre-training on Kinetics 400~\cite{6kinetic400} and hybrid dataset}, where we initialized the backbone with Kinetics 400 or Hybrid dataset pre-trained weights and fine-tuned the entire model on the FHA-Kitchens training set; and (2) \textbf{Training from scratch on FHA-Kitchens}, where we randomly initialized the model weights and directly train them on FHA-Kitchens. For different models, we used the recommended optimization strategy and batch size, and the maximum training period was set to 210 epochs.

\begin{table*}[ht!]
  \centering
  \caption{Classification results (Top-1 and Top-5 accuracy) of fine-grained hand actions using different features and the skeleton-based STGCN~\cite{stgcn} method (pre-trained on NTU60~\cite{ntu60} and NTU120~\cite{ntu120}) on the validation set of the SL-AR track.}
  \begin{tabular}{cccc|cccc}
  \toprule
  Pre-train Data & Feature & Top-1 & Top-5 & Pre-train Data & Feature & Top-1 & Top-5 \\
  \midrule
 \multirow{8}{*}{NTU60}  & joint-2d & 22.78 & 47.68 & \multirow{8}{*}{NTU120} & joint-2d & 20.68 & 48.10\\
 \multirow{8}{*}{~} & joint-3d & 22.36 & 52.32 & \multirow{8}{*}{~} & joint-3d & 21.10 & 47.68\\
  \multirow{8}{*}{~} & joint-motion-2d & 8.02 & 19.83 & \multirow{8}{*}{~} & joint-motion-2d & 9.28 & 20.25\\
  \multirow{8}{*}{~} & joint-motion-3d & 10.97 & 23.63 & \multirow{8}{*}{~} & joint-motion-3d & 11.81 & 26.16\\
 \multirow{8}{*}{~} & bone-2d & 22.36 & 49.79 & \multirow{8}{*}{~} & bone-2d & 24.05 & 57.81\\
  \multirow{8}{*}{~} & bone-3d & 24.05 & 52.32 & \multirow{8}{*}{~} & bone-3d & 24.05 &  51.05\\
   {\multirow{8}{*}{~}} &  bone-motion-2d &  10.55 &  23.21 &  \multirow{8}{*}{~} &  bone-motion-2d &  9.28 &  23.21\\
   \multirow{8}{*}{~} &  bone-motion-3d &  13.50 &  26.16 &  \multirow{8}{*}{~} & bone-motion-3d & 12.24 & 27.00\\
  \bottomrule
 \end{tabular}
  \label{sup_skeletonresults}%
\end{table*}%

 \textbf{Results on the SL-AR Track.} Table~\ref{arfullresults} presents the performance of different action recognition methods on the Kinetics 400~\cite{6kinetic400} dataset and the proposed FHA-Kitchens dataset, with and without pre-trained models. From the experimental results, it can be observed that the performance trends of all action recognition methods on FHA-Kitchens are similar to their performance on Kinetics 400~\cite{6kinetic400}, while the models perform much better on the coarse-grained actions of Kinetics 400. For the best-performing VideoSwin~\cite{23videoswin} model, the top-1 accuracy on Kinetics 400 surpasses the top-1 accuracy on FHA-Kitchens by 43.44\%. And those methods with even large models cannot achieve satisfactory performance. This is clear evidence that validates the challenging nature of the fine-grained hand action recognition on FHA-Kitchens. Besides, the utilization of pre-trained weights has proven to be beneficial, resulting in improved accuracy compared to training models from scratch. This finding suggests that despite the existence of a domain gap between coarse-grained and fine-grained actions, pre-training remains an effective strategy for addressing the challenges inherent in FHA-Kitchens, which have a larger number of action categories and relatively limited training data.

 In addition, we further supplemented the hand pose information and conducted experiments using the skeleton-based STGCN~\cite{stgcn} method. We used STGCN pre-trained on NTU60~\cite{ntu60} and NTU120~\cite{ntu120} and fine-tuned the models on the SL-AR track using different features for fine-grained hand actions, the results (Top-1 and Top-5 accuracy) can be seen in Table~\ref{sup_skeletonresults}.

 According to the experimental results, it can be observed that 3D pose features outperform 2D pose features and bone features achieve better results than joint features. Nevertheless, the overall results did not surpass the efficacy of hand-object interaction-based approaches, highlighting that relying only on hand pose information is insufficient for accomplishing fine-grained action recognition tasks. Because the generation of hand actions involves interacting objects, achieving a fine-grained hand action recognition task is required to consider the information of the objects interacting with the hand, which is different from a whole-body action recognition task (e.g., AVA, FineGym dataset).

\textbf{DG Track: Intra- and Inter-class Domain Generalization for Interaction Region Detection}

\begin{itemize}[leftmargin=*]
\item \textbf{\emph{Intra-class Domain Generalization}}~\label{intra}

 \begin{table}[ht]
    \caption{Intra-class DG test results of Faster RCNN~\cite{24faster} with the ResNet50 backbone on the ``\textbf{Cut}'' Setting. $\Delta_i = \frac{1}{3}\sum_{j, j\neq i}{ji} - ii, \Delta^{*}_i=\frac{1}{3}\sum_{j, j\neq i}{ij} - ii, i=0,1,2,3$.}
    \label{dgtrack-cut}
    \centering
    \setlength{\tabcolsep}{1mm}{
    \begin{tabular}{c|cccc|c}
    \toprule
     & \multicolumn{4}{c}{Test (mAP)} & \\
    \cmidrule(r){2-5}
     Train & cut-slice & cut-off & cut-down & cut-dice & $\Delta^{*}$\\
    \midrule
    w/o cut-slice & \textbf{33.30} & 65.00 & 56.00 & 60.90 & 27.33\\
    w/o cut-off & 57.10 & \textbf{48.00} & 54.80 & 62.80 & 10.23\\
    w/o cut-down & 57.30 & 64.40 & \textbf{41.30} & 63.50 & 20.43\\
    w/o cut-dice & 57.50 & 64.90 & 58.70 & \textbf{41.10} & 19.27 \\
    \midrule
    $\Delta$ & 24.00 & 16.77 & 15.20 & 21.30 &  \\
    \bottomrule
    \end{tabular}}
\end{table}

\begin{table}[ht]
  \centering
  \caption{Intra-class DG test results of Faster RCNN~\cite{24faster} with the ResNet50 backbone on the ``{\bf Hold}'' Setting. $\Delta_i = \frac{1}{2}\sum_{j, j\neq i}{ji} - ii, \Delta^{*}_i=\frac{1}{2}\sum_{j, j\neq i}{ij} - ii, i=0,1,2$.}
  \label{dgtrack-hold}
    \centering
    \begin{tabular}{c|ccc|c}
    \toprule
     & \multicolumn{3}{c}{Test (mAP)} & \\
    \cmidrule(r){2-4}
     Train & hold-up & hold-in & hold-around & $\Delta^{*}$\\
    \midrule
    w/o hold-up & \textbf{44.00} & 53.50 & 71.70 & 18.60   \\
    w/o hold-in & 44.30 & \textbf{7.30} & 69.10 & 49.40\\
    w/o hold-around & 52.30 & 52.80 & \textbf{47.80} & 4.75 \\
    \midrule
    $\Delta$ & 4.30 & 45.85 & 22.60 &  \\
    \bottomrule
   \end{tabular}
\end{table}

\begin{table}[ht]
  \centering
  \caption{Intra-class DG test results of Faster RCNN~\cite{24faster} with the ResNet50 backbone on the ``{\bf Take}'' Setting. $\Delta_i = \frac{1}{2}\sum_{j, j\neq i}{ji} - ii, \Delta^{*}_i=\frac{1}{2}\sum_{j, j\neq i}{ij} - ii, i=0,1,2$.}
  \label{dgtrack-take}
    \centering
    \begin{tabular}{c|ccc|c}
    \toprule
     & \multicolumn{3}{c}{Test (mAP)} & \\
    \cmidrule(r){2-4}
     Train & pick-up & grab & catch & $\Delta^{*}$\\
    \midrule
    w/o pick-up & \textbf{0.40} & 47.10 & 46.60 & 46.45\\
    w/o grab & 19.00 & \textbf{4.50} & 39.00 & 24.50\\
    w/o catch & 19.10 & 46.00 & \textbf{15.60} & 16.95 \\
    \midrule
    $\Delta$ & 18.65 & 42.05 & 27.20 &  \\
    \bottomrule
   \end{tabular}
\end{table}

 \textbf{Settings.} We conducted intra-class DG experiments using the three most prevalent parent action categories, \textit{i.e.}, \textit{Cut}, \textit{Hold}, and \textit{Take}. For each parent action category, we selected the most prevalent sub-categories and adopted the cross-validation protocol, \textit{i.e.}, randomly choosing one sub-category as the test set while using all other sub-categories for training. Following the SL-AD track, we selected the Faster RCNN~\cite{24faster} model with the ResNet50 backbone as the default model, which is pre-trained on the MS COCO~\cite{53coco} object detection dataset.

 \textbf{Results on the Intra-class DG Track.} The results on \textit{Cut}, \textit{Hold}, and \textit{Take} are summarized in Table~\ref{dgtrack-cut}, ~\ref{dgtrack-hold}, and ~\ref{dgtrack-take}. In the ``\textit{Cut}'' parent category, the performance of all four detection models remains stable for the sub-categories seen during training but deteriorates for unseen sub-categories, as evidenced by the diagonal scores, which exhibit a minimum drop of 15 mAP. The findings in the results of the other two parent categories align with the observations in the ``\textit{Cut}'' parent category. This finding suggests that there is still potential for enhancing the models' generalization abilities, \textit{e.g.}, by exploring the domain generalization or unsupervised domain adaptation techniques.

\item \textbf{\emph{Inter-class Domain Generalization}}

 \begin{table}[b]
    \caption{Inter-class DG test results. $\Delta_i=ii-\frac{1}{2}\sum_{j,j\neq i}{ji}, \Delta^{*}_i=ii-\frac{1}{2}\sum_{j, j\neq i}{ij}, i=0,1,2$.}
    \label{dgtrack-inter}
    \centering
    \setlength{\tabcolsep}{1.8mm}{\begin{tabular}{c|ccc|c}
    \toprule
     & \multicolumn{3}{c}{Test (mAP)} & \\
     \cmidrule(r){2-4}
    Train & Cut & Hold & Take & $\Delta^{*}$\\
    \midrule
    Cut & \textbf{37.40} & 29.50 & 29.20 & 8.05\\
    Hold & 48.70 & \textbf{52.30} & 41.80 & 7.05\\
    Take & 14.00 & 13.20 & \textbf{41.20} & 27.60\\
    \midrule
    $\Delta$ & 6.05 & 30.95 & 5.70\\
    \bottomrule
    \end{tabular}}
\end{table}

\textbf{Settings.} We chose the three most prevalent parent action categories \textit{Cut}, \textit{Hold}, and \textit{Take}, and adopted the cross-validation protocol, \textit{i.e.}, randomly choosing one parent category for training and using the other parent categories for testing. Other settings follow those in the intra-class DG track.

\textbf{Results on the Inter-class DG Track.} The results are listed in Table \ref{dgtrack-inter}. Similar to the results in the intra-class DG track, the detection models perform well on the seen categories while deteriorating on the unseen categories. Nevertheless, it is interesting to find that the performance gap ($\bigtriangleup_0=6.05$ and $\bigtriangleup^{*}_0=8.05$) between \textit{Cut} and others are smaller than those in the intra-class DG track, implying that there is likely a large intra-class variance, and the detection model is prone to overfitting the seen categories, particularly when the volume of training data is smaller (there are 7,463 training frames in \textit{Hold} while only 1,680 in \textit{Take}).

\end{itemize}

\textbf{2) Qualitative Results}~\label{qualitative}
\

\begin{figure*}[htb]
  \centering
   \includegraphics[width=0.93\linewidth]{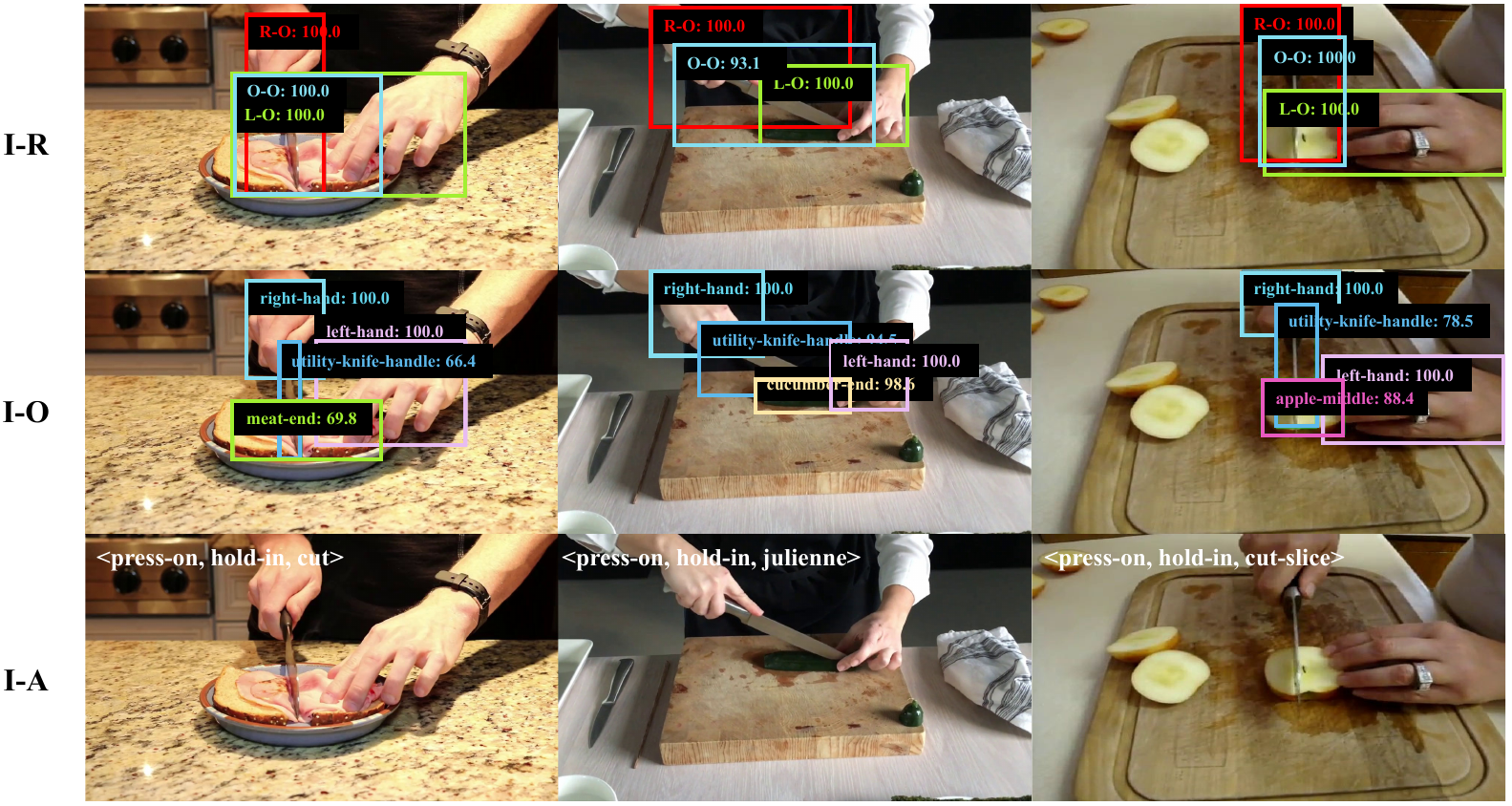}
  \caption{Visual results of Faster-RCNN~\cite{24faster} and TSN~\cite{19TSN} methods in the SL-AD and SL-AR track experiments on our FHA-Kitchens dataset, showcasing interaction scenes with \emph{three} hand sub-interaction regions, \textit{i.e.}, ``Left hand-Object interaction region (L-O)'', ``Right hand-Object interaction region (R-O)'', and ``Object-Object interaction region (O-O)''. I-R: Interaction Region, I-O: Interaction Object, I-A: Interaction Region Action Verb.} 
  \label{datasetvis3}
\end{figure*}

\begin{figure*}[htb]
  \centering
  \includegraphics[width=0.94\linewidth]{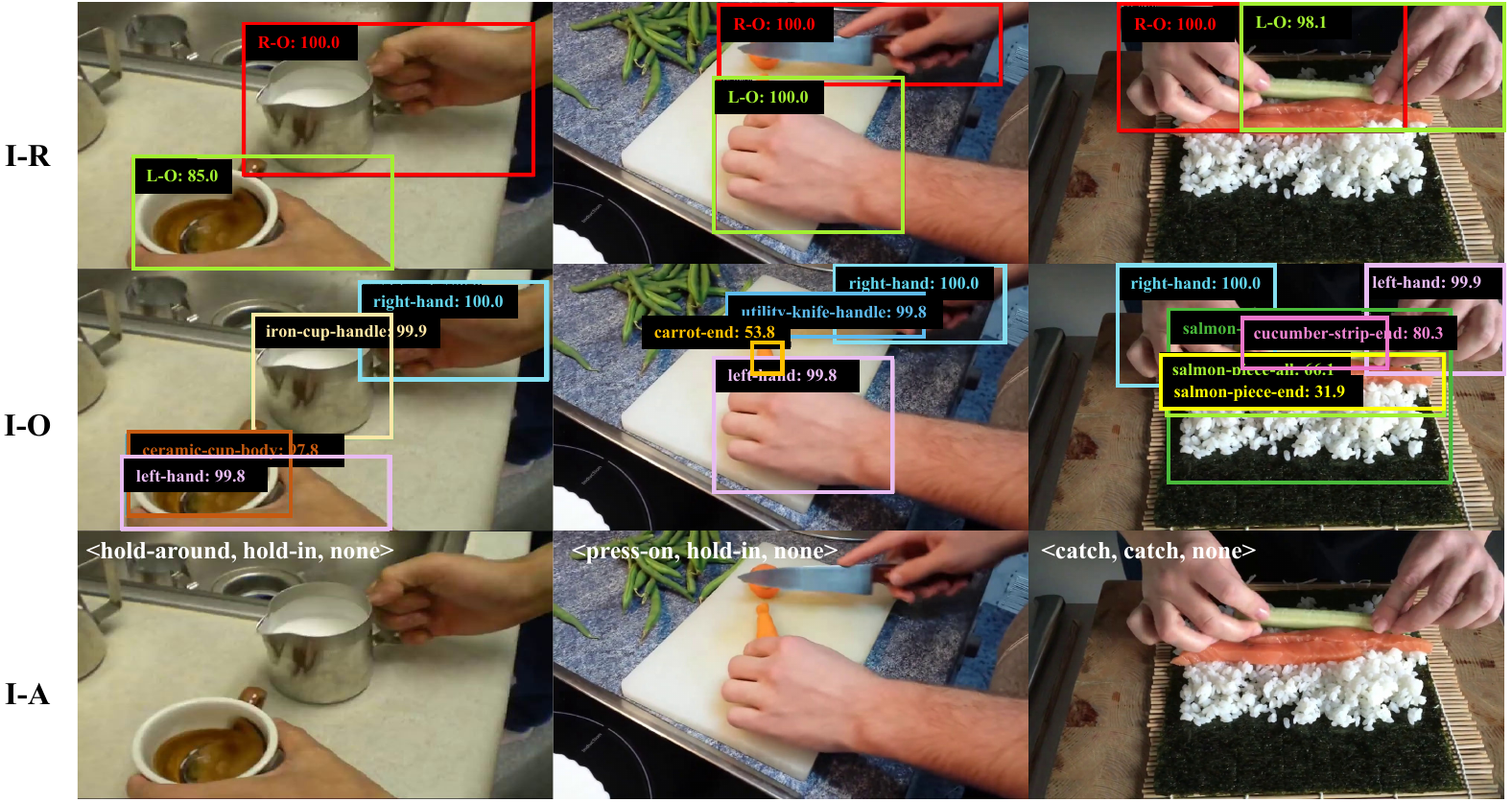}
  \caption{Visual results of Faster-RCNN~\cite{24faster} and TSN~\cite{19TSN} methods in the SL-AD and SL-AR track experiments on our FHA-Kitchens dataset, showcasing interaction scenes with \emph{two} hand sub-interaction regions, \textit{i.e.}, ``Left hand-Object interaction region (L-O)'' and ``Right hand-Object interaction region (R-O)''. I-R: Interaction Region, I-O: Interaction Object, I-A: Interaction Region Action Verb.} 
  \label{datasetvis2}
\end{figure*}

\begin{figure*}[htb]
  \centering
  \includegraphics[width=0.95\linewidth]{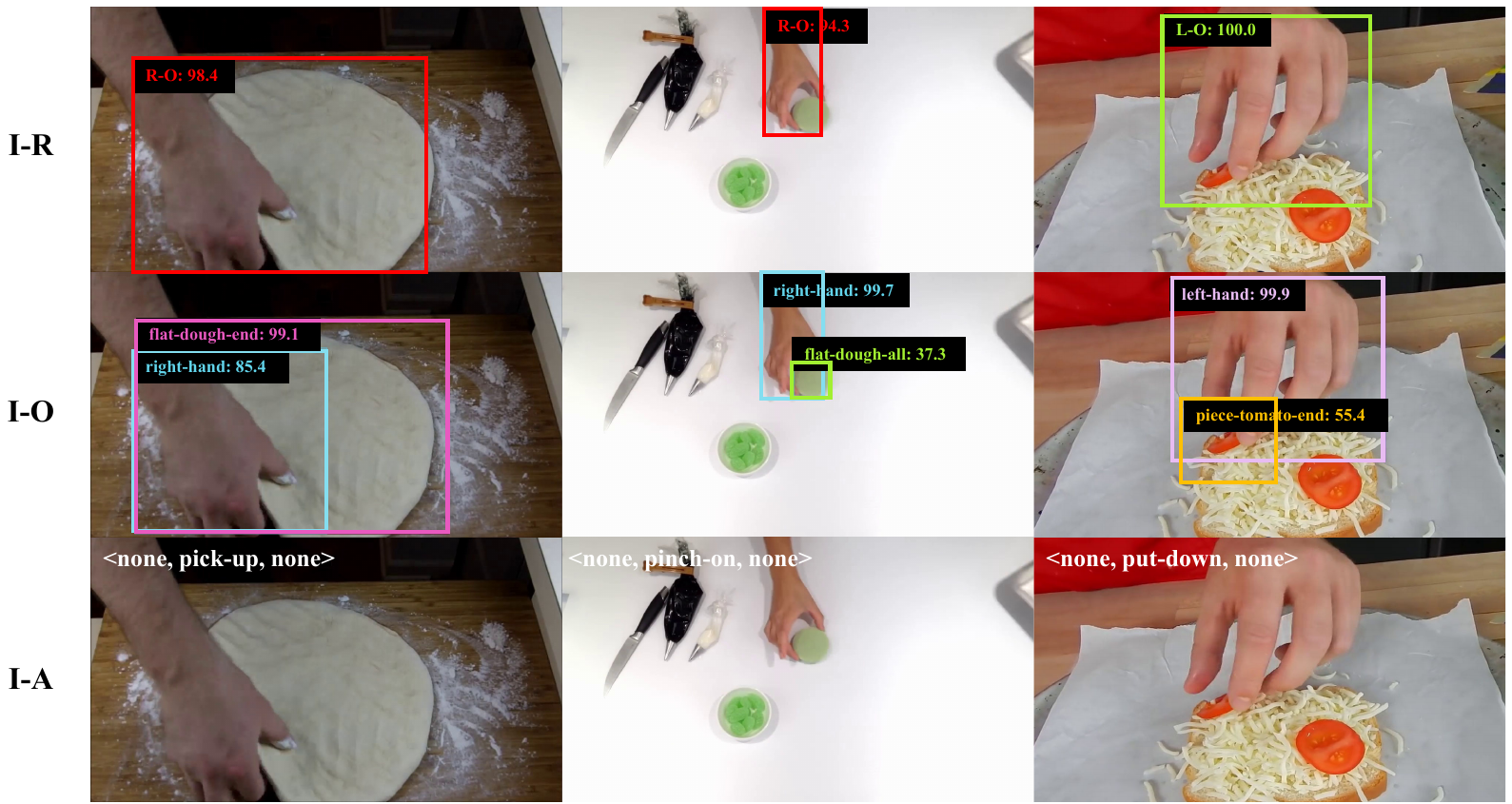}
  \caption{Visual results of Faster-RCNN~\cite{24faster} and TSN~\cite{19TSN} methods in the SL-AD and SL-AR track experiments on our FHA-Kitchens dataset, showcasing interaction scenes with a \emph{single} hand sub-interaction region, \textit{i.e.}, ``Left hand-Object interaction region (L-O)'' or ``Right hand-Object interaction region (R-O)''. I-R: Interaction Region, I-O: Interaction Object, I-A: Interaction Region Action Verb.} 
  \label{datasetvis1}
\end{figure*}

The visual results of the SL-AD and SL-AR track experiments are presented in Figure~\ref{datasetvis3}, Figure~\ref{datasetvis2}, and Figure~\ref{datasetvis1}. We showcased the visualization results of interaction region detection, interacting object detection, and action recognition, focusing on hand interaction scenarios of varying complexity. In the interaction region detection results, we provide coarse-grained action categories corresponding to the sub-interaction regions, \textit{i.e.}, \emph{<L-O, R-O, O-O>}. In the recognition results, we provide fine-grained action verbs corresponding to the three hand sub-interaction regions, denoted as \emph{<L-O action verb, R-O action verb, O-O action verb>}. Figure~\ref{datasetvis3} shows some challenging cases of hand interactions, providing compelling evidence of the good prediction performance of detection and recognition models, \textit{i.e.}, the Faster-RCNN~\cite{24faster} with a ResNet50 backbone for detection and a pre-trained TSN~\cite{19TSN} model with a ResNet50 backbone for action recognition. Moreover, Figure~\ref{datasetvis2} and Figure~\ref{datasetvis1} also demonstrate accurate detection and recognition results for some common interaction cases.

 \begin{table}[t!]
  \caption{Comparative experiments of the action dimensional weight, \textit{i.e.}, hyper-parameter $w_d$ ($d$ $\in$ \{s, a, o\}), under different ratio settings. $w_s = w_o = (1-w_a)/2$, $w_s + w_a + w_o = 1$, M-G: Mixed-Grained.}
      \centering
      \setlength{\tabcolsep}{1mm}{
      \begin{tabular}{cc|c|cc}
      \toprule
           Method & Backbone & Train Data & $w_a$ & M-G mAP(\%)  \\
       \midrule
           \multirow{5}{*}{\textbf{\emph{Ours-4scale}}} &  \multirow{5}{*}{ResNet-50} & \multirow{11}{*}{Multi-Granularity} & 0.5 & 54.9\\
           \multirow{5}{*}{~} & \multirow{5}{*}{~} & \multirow{11}{*}{~} & \textbf{0.6} & \textbf{57.0}\\
           \multirow{5}{*}{~}& \multirow{5}{*}{~} & \multirow{11}{*}{~} & 0.7 & 55.6\\
           \multirow{5}{*}{~} & \multirow{5}{*}{~} & \multirow{11}{*}{~}& 0.8 & 56.0\\
           \multirow{5}{*}{~} & \multirow{5}{*}{~} & \multirow{11}{*}{~}& 0.9 & 54.7\\
           \cmidrule(r){1-2}
           \cmidrule(r){4-5}
           &&&&\\
           \cmidrule(r){1-2}
           \cmidrule(r){4-5}
           \multirow{5}{*}{\textbf{\emph{Ours-5scale}}} &  \multirow{5}{*}{Swin-L} & \multirow{11}{*}{~} & 0.5 & 57.2\\
           \multirow{5}{*}{~} & \multirow{5}{*}{~} & \multirow{11}{*}{~} & \textbf{0.6} & \textbf{59.4}\\
           \multirow{5}{*}{~}& \multirow{5}{*}{~} & \multirow{11}{*}{~} & 0.7 & 58.2\\
           \multirow{5}{*}{~} & \multirow{5}{*}{~} & \multirow{11}{*}{~}& 0.8 & 57.4\\
           \multirow{5}{*}{~} & \multirow{5}{*}{~} & \multirow{11}{*}{~}& 0.9 & 57.1\\
        \bottomrule
      \end{tabular}}
      \label{tab:weight}
  \end{table}

\subsection{MG-HAD: Multi-Granularity Hand Action Detection}

\subsubsection{\textbf{M}ulti-\textbf{d}imensional \textbf{A}ction \textbf{Q}ueries}
\

 To enhance the model's understanding of multi-dimensional action information from global and local perspectives, in our design of multi-dimensional action queries, we introduce an action dimensional hyper-parameter $w_d$ ($d$ $\in$ \{s, a, o\}), to control the proportion of local information (sub-categories) fused with global information (triplet categories). In the three action dimensions \emph{<s, a, o>}, the \emph{a} dimension is the most crucial for our task. Therefore, we use $w_a$ as the key weight to dynamically adjust the weight proportions of the three action dimensions, with a total sum of 1. To determine the optimal weight distribution, we conducted a total of 10 comparative experiments under different backbones, with detailed results provided in Table~\ref{tab:weight}. Based on the experimental results, the action dimensional weight setting we selected is as follows:
\begin{equation}
      w\_d = \left\{
    \begin{array}{ll}
    0.6 & \text{if } d = a \\
    0.2 & \text{if } d = s \text{ or } d = o
    \end{array}
    \right.,
\end{equation}
where $w_s = w_o = (1-w_a)/2$, $w_s + w_a + w_o = 1$.

\begin{table*}[t!]
  \caption{Three sets comparative experiments of the ``noise label generation strategy'' in the \textbf{C}oarse-\text{F}ine \textbf{C}ontrastive \textbf{D}e\textbf{N}oising (C-F CDN). M-G: Mixed-Grained, C-G: Coarse-Grained, F-G: Fine-Grained.} 
      \centering  
      \begin{tabular}{c|cc|c|c|ccc}
      \toprule
           \multirow{2}{*}{Method} &\multirow{2}{*}{Backbone} & \multirow{2}{*}{Train Data} & \multirow{2}{*}{noisy label location} & \multirow{2}{*}{noisy label class} & \multicolumn{3}{c}{FHA-Kitchens val mAP(\%)} \\
           \cmidrule(r){6-8}
           \multirow{2}{*}{~} &\multirow{2}{*}{~} & \multirow{2}{*}{~} & \multirow{2}{*}{~}& \multirow{2}{*}{~} & M-G label & C-G sub-label & F-G sub-label\\
       \midrule
          \emph{Baseline}~\cite{dino} &  \multirow{8}{*}{ResNet-50} & \multirow{8}{*}{multi-granularity} & random & random & 54.7 & 76.3 & 54.5\\
        \cline{1-1}
        \cline{4-8}
          \multirow{6}{*}{\textbf{\emph{Ours-4scale}}} & \multirow{8}{*}{~} & \multirow{8}{*}{~}& & $a_1 \rightarrow$ mixed-grained & \multirow{2}{*}{\textbf{56.6}} & \multirow{2}{*}{75.9} & \multirow{2}{*}{\textbf{56.4}}\\
          \multirow{6}{*}{~} & \multirow{8}{*}{~} & \multirow{8}{*}{~}& & $a_2 \rightarrow$ fine-grained & \multirow{2}{*}{~} & \multirow{2}{*}{~} &\multirow{2}{*}{~}\\
        \cline{5-8}
          \multirow{6}{*}{~} & \multirow{8}{*}{~} & \multirow{8}{*}{~}& $a_1$: fine-grained & $a_1 \rightarrow$ coarse-grained & \multirow{2}{*}{56.1} & \multirow{2}{*}{75.5} & \multirow{2}{*}{55.8}\\
          \multirow{6}{*}{~} & \multirow{8}{*}{~} & \multirow{8}{*}{~}& $a_2$: coarse-grained & $a_2 \rightarrow$ fine-grained & \multirow{2}{*}{~} & \multirow{2}{*}{~} &\multirow{2}{*}{~}\\
        \cline{5-8}
          \multirow{6}{*}{~} & \multirow{8}{*}{~} & \multirow{8}{*}{~} &  & $a_1 \rightarrow$ coarse-grained & \multirow{2}{*}{55.8} & \multirow{2}{*}{74.5} &\multirow{2}{*}{55.6}\\
          \multirow{6}{*}{~} & \multirow{8}{*}{~} & \multirow{8}{*}{~}& & $a_2 \rightarrow$ mixed-grained & \multirow{2}{*}{~} & \multirow{2}{*}{~} &\multirow{2}{*}{~}\\
        
        \bottomrule
      \end{tabular}
      \label{tab:cdn}
  \end{table*}

\subsubsection{\textbf{C}oarse-\text{F}ine \textbf{C}ontrastive \textbf{D}e\textbf{N}oising (C-F CDN)}
\

To enable the model to handle multi-granularity hand action labels and understand the differences between different granularity labels, we propose the C-F CDN module. When designing the noise label generation strategy, we replaced the ``random'' generation of noise with ``specified'', reducing randomness by specifying noise positions and categories. This ensures noise is added to different granularity labels, generating CDN queries encompassing various granularity. To determine the optimal setting, we considered the noise distribution of different granularity categories in real-world scenarios and ensured contrastive learning between coarse- and fine-grained information. We conducted three sets of comparative experiments (see Table~\ref{tab:cdn}). The final selected setting, as shown in Eq.~(\ref{equ:c-f}), exhibits the most significant improvement. Hence, subsequent experiments were conducted using this setting for further investigation. Our method's success lies in its ability to suppress confusion at the category level and select appropriate granularity to predict hand action categories, thus enhancing its ability to predict multi-granularity information.
    \begin{equation}\label{equ:c-f}
    \text{noise label} = \left\{
    \begin{array}{ll}
    \text{fine-grained} & i \in [0,3) \\
    \text{mixed-grained} & i \in [3,C)
    \end{array}
    \right.,
    \end{equation}
where $i$ indexes the multi-granularity action category for a specific instance (\textit{i.e.}, 0$\sim$2 denote coarse-grained categories while 3$\sim$C-1 denote fine-grained categories), and $C$ is the number of categories. ``fine-grained'' and ``mixed-grained'' denote that the noise label is chosen randomly from the fine-grained categories and the combination of the coarse-grained and fine-grained categories, respectively.



\subsubsection{Action Detection Results of MG-HAD}

\

\begin{figure*}[ht!]
 \centering
 \includegraphics[width=1\linewidth]{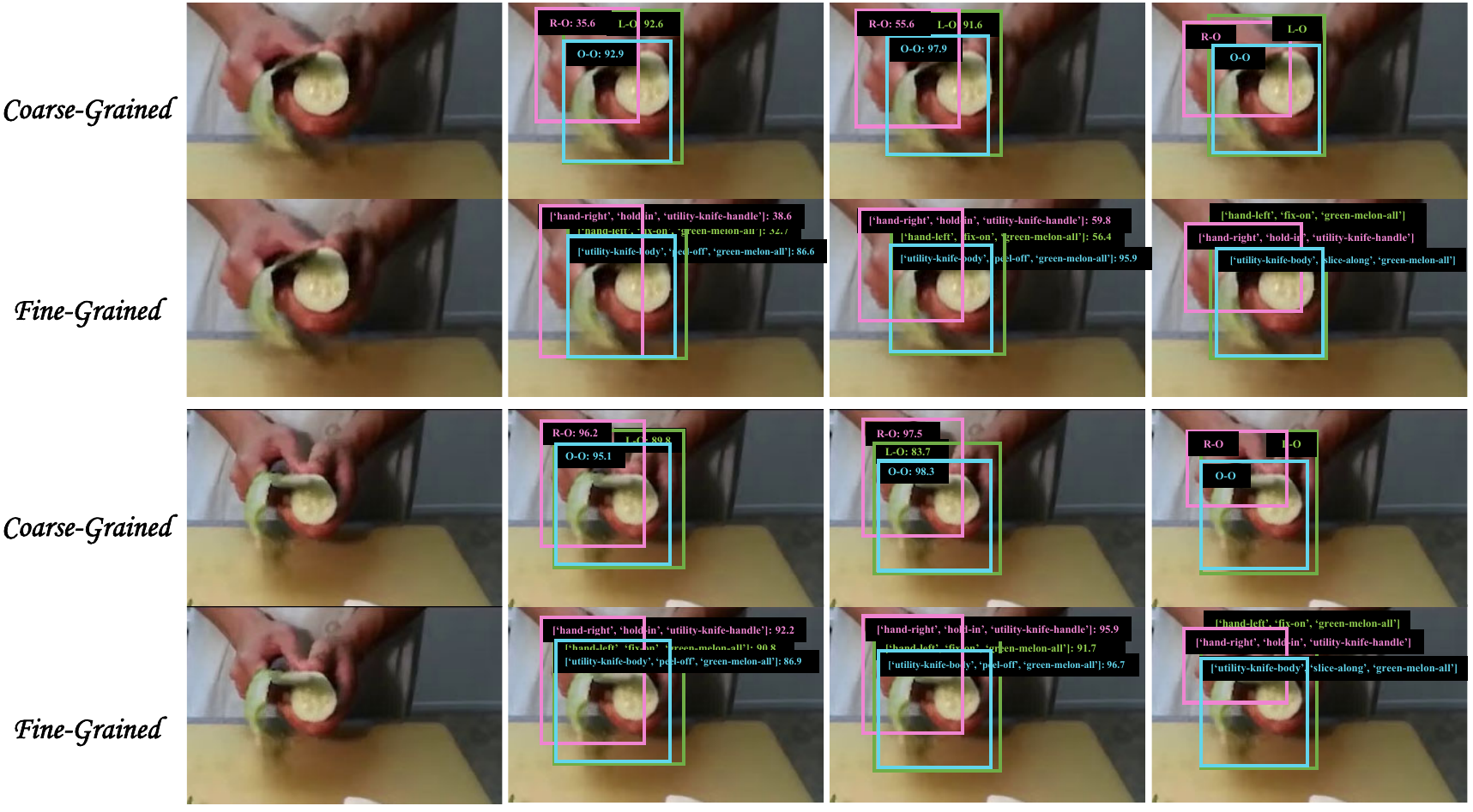} \\
 \includegraphics[width=1\linewidth]{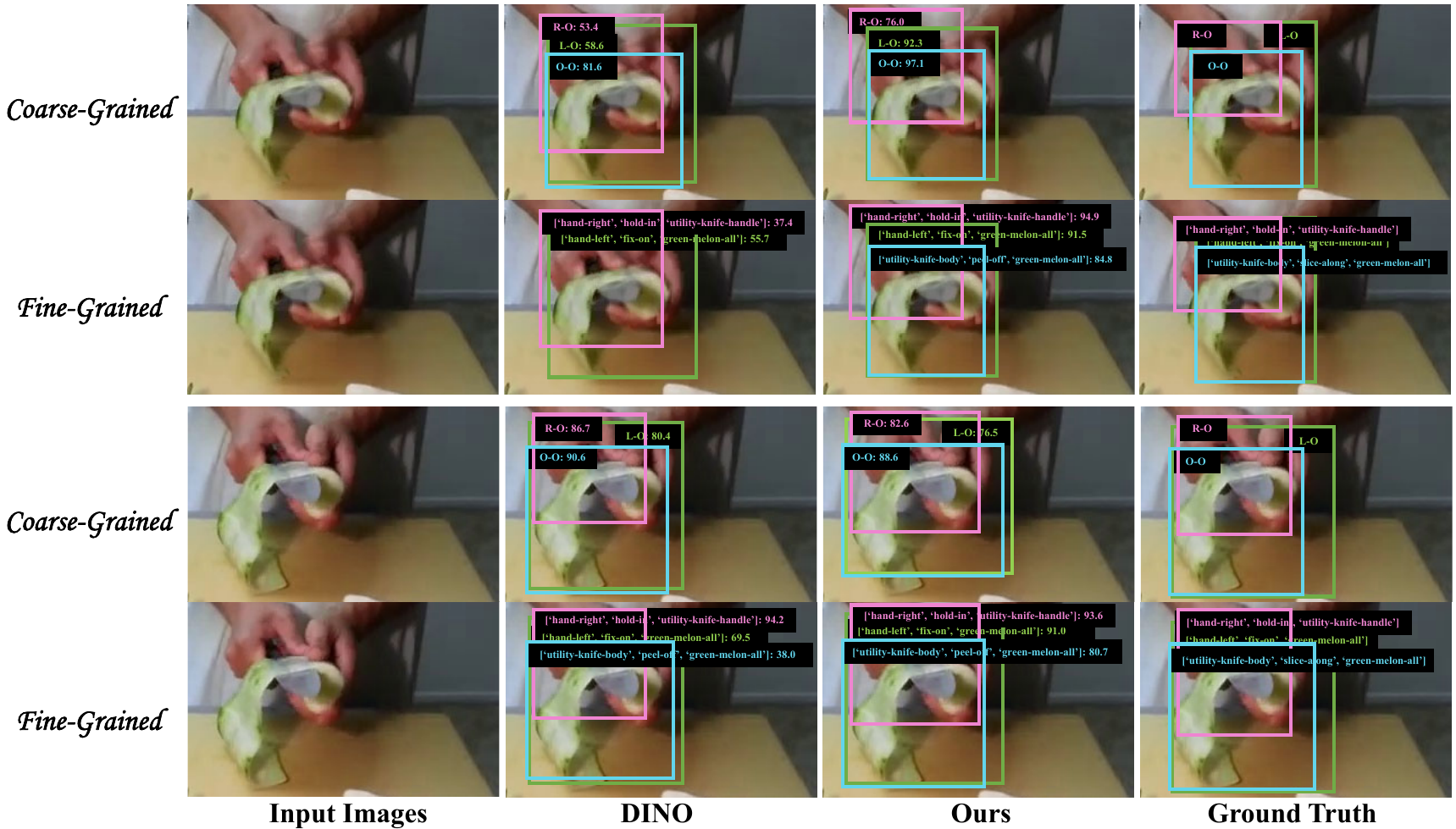}
 \caption{Qualitative comparison on the FHA-Kitchens dataset. Our model accurately detects three hand sub-interaction regions and provides multi-granularity hand action categories. Compared to the baseline~\cite{dino}, our model performs better across multi-dimensional fine-grained categories, demonstrating the effectiveness of our designed multi-dimensional action queries.}
 \label{fig:comparevis}
\end{figure*}

\begin{figure*}[ht!]
    \centering  \includegraphics[width=0.9\linewidth]{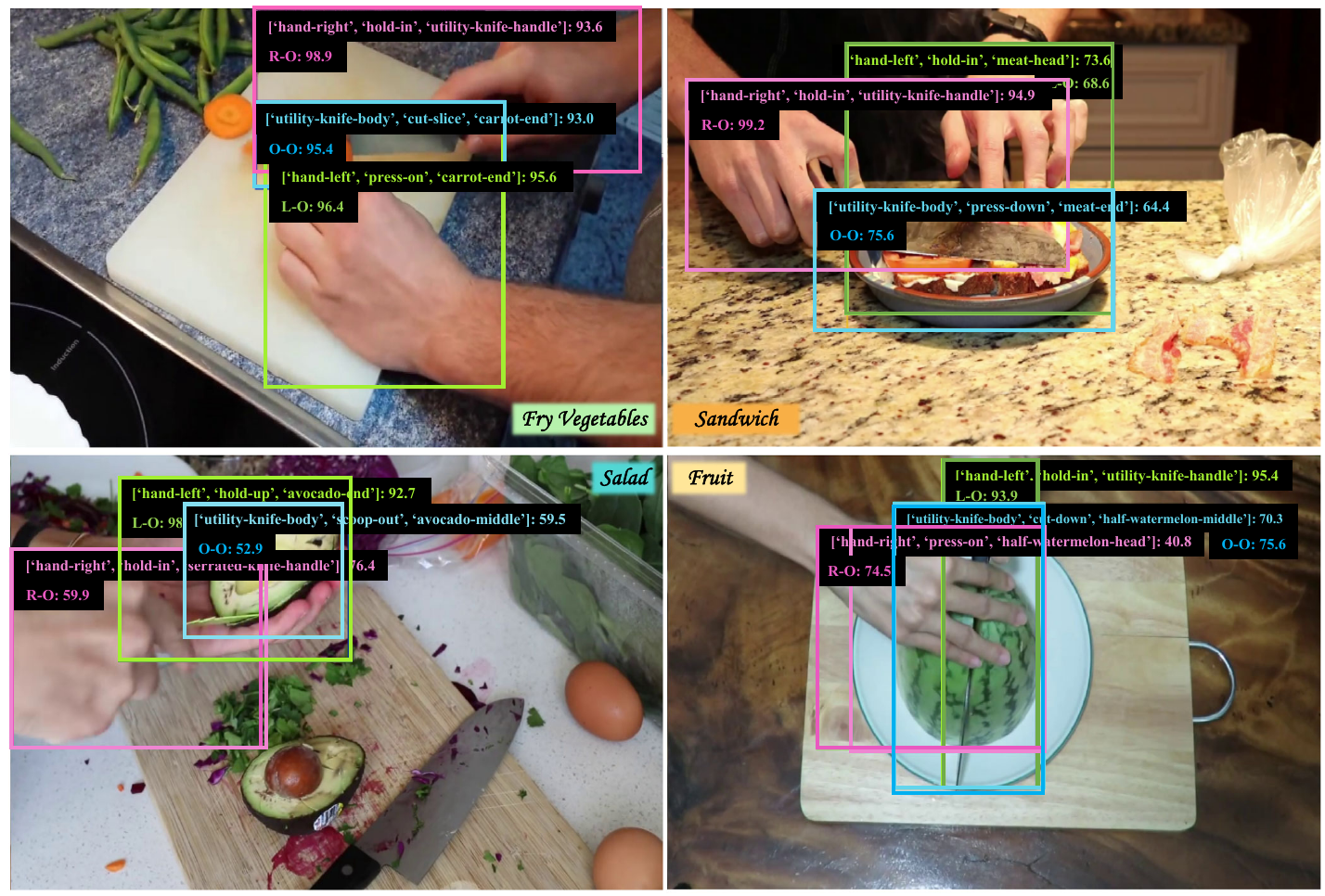}
    \caption{Visual detection results of our method in \emph{four} different kitchens scenarios containing complex hand actions, \textit{i.e.}, ``\emph{fry vegetables}'', ``\emph{sandwich}'', ``\emph{salad}'', and ``\emph{fruit}''. 
    Our model offers accurate bounding boxes and multi-granularity hand action information for three hand sub-interaction regions.
    } \label{fig:ourvisdet}
\end{figure*}

 \textbf{Visualization detection results.} Based on the 5scale-Swin-L backbone, qualitative comparison results with the baseline~\cite{dino} on the FHA-Kitchens dataset are shown in Figure~\ref{fig:comparevis}. Our model accurately detects three hand sub-interaction regions (\textit{i.e.}, ``Left hand-Object interaction region (L-O)'', ``Right hand-Object interaction region (R-O)'', and ``Object-Object interaction region (O-O)'') and provides multi-granularity hand action categories (\textit{i.e.}, ``Coarse-Grained'' and ``Fine-Grained''). Compared to the DINO, we demonstrate superior performance across multiple dimensions of fine-grained categories, illustrating the effectiveness of our designed multi-dimensional action queries. Additionally, we present our model's multi-granularity hand action detection results in more kitchen scenarios, as shown in Figure~\ref{fig:ourvisdet}. We randomly selected four different kitchen scenarios, \textit{i.e.}, ``\emph{fry vegetables}'', ``\emph{sandwich}'', ``\emph{salad}'', and ``\emph{fruit}'', showcasing complex hand actions. Our model offers accurate bounding boxes and multi-granularity hand action information for three hand sub-interaction regions.

 \textbf{Basic Hyper-parameters.} For the basic hyper-parameters, consistent with DINO~\cite{dino}, we utilized a 6-layer Transformer encoder and a 6-layer Transformer decoder with a hidden feature dimension of 256. We set the initial learning rate (lr) to ${1 \times 10^{-4}}$ and employed a MultiStep lr scheduler, dropping lr by multiplying 0.1 at the 11-th and 20-th epochs for ResNet50, corresponding to the 12 and 24 epoch settings. We employed the AdamW~\cite{adam,weightdecay} optimizer with a weight decay of ${1 \times 10^{-4}}$ and trained our model on NVIDIA GeForce RTX 3090 GPUs. The ResNet50 backbone was trained with a batch size of 2 per GPU, while the SwinL backbone had a batch size of 1. We initialized 900 decoder queries, maintaining the same computational cost as DINO. We provide detailed hyper-parameters in Table~\ref{tab:hp} for reproducibility.

\subsection{Datasheets for Datasets}
\subsubsection{Motivation}
\ 

\noindent \textbf{1. For what purpose was the dataset created? Was there a specific task in mind? Was there a specific gap that needed to be filled? Please provide a description.}

\textbf{A1:} FHA-Kitchens is created to facilitate research in the field of complex multi-granularity hand action. It is important to study several challenging questions in the context of more training data from diverse multi-granularity hand actions, such as: (1) How do different representative action recognition models perform on fine-grained hand action tasks? (2) How do state-of-the-art detection models perform on the refined hand interaction regions with multi-granularity hand action categories? (3) How about the impact of pre-training, e.g., on the whole-body actions dataset~\cite{6kinetic400}, in the context of the large-scale dataset with diverse multi-granularity hand actions? and (4) How do the intra-class and inter-class generalization capabilities of models trained with specific fine-grained hand actions or parent hand actions perform? However, existing action datasets primarily focus on whole-body actions or coarse-grained action categories, lacking finer-grained hand-action localization and category information. Therefore, it is impossible to study these questions using existing datasets. In contrast, FHA-Kitchens primarily focuses on hand actions and refines hand interaction regions into three sub-interaction regions. We annotated coarse- and fine-grained actions for each sub-interaction region. Coarse-grained categories, denoted by the generic terms ``L-O'', ``R-O'', and ``O-O'', represent the coarse actions within the sub-interaction regions. Fine-grained action category in a triplet format: \emph{<subject, action verb, object>}. Overall, we meticulously annotated 880 hand action categories (coarse- and fine-grained) for approximately 220k bounding boxes, with each category corresponding to a sub-interaction region’s localization box. Fine-grained categories per frame have nine dimensions, resulting in 877 action triplets, significantly enhancing the granularity of actions and providing valuable resources for researchers to study these questions effectively.

FHA-Kitchens aims to provide a better, more comprehensive, and finer-grained benchmark for hand action. However, existing hand-action datasets exhibit limitations including insufficient representation of hand-action granularity, lack of annotation of hand-action interaction regions, and neglect of the relationships between interacting objects. With its diverse and finer-grained hand action information, the FHA-Kitchens dataset enables a better evaluation performance for hand action tasks.

\noindent \textbf{2. Who created this dataset (e.g., which team, research group) and on behalf of which entity (e.g., company, institution, organization)?}

\textbf{A2:} Our dataset is created by the authors as well as some volunteer undergraduate students from Wuhan University, including Ting Zhe, Yongqian Li, Chengli Chen, Xin Ding, Jiali Li, Tengzheng Li, RunFu Guo, Heng Gao, Xing Zhao, Weisheng Chen, Chaoyu Mai, Yipan Wei.

\noindent \textbf{3. Who funded the creation of the dataset? If there is an associated grant, please provide the name of the grantor and the grant name and number.}

\textbf{A3:} This work was supported in part by the National Key Research and Development Program of China under No. 2021YFC33002 00, the National Natural Science Foundation of China (Grant No. 62276195), and the Special Fund of Hubei Luojia Laboratory under Grant 220100014.

\subsubsection{Composition}~\label{A3.2}

\noindent \textbf{1. What do the instances that comprise the dataset represent (e.g., documents, photos, people, countries)? Are there multiple types of instances(e.g., movies, users, and ratings; people and interactions between them; nodes and edges)? Please provide a description.}

\textbf{A1:} FHA-Kitchens consists of video clips, each video clip consists of consecutive video frames, including 880 coarse- and fine-grained hand action categories. For each frame, we provide bounding boxes for three hand sub-interaction regions (\textit{i.e.}, left hand-object (L-O), right hand-object (R-O), and object-object (O-O) interaction regions) and the interaction objects. Each sub-interaction region action was annotated using coarse- and fine-grained action categories. Coarse-grained categories, denoted by the generic terms ``L-O'', ``R-O'', and ``O-O'', fine-grained action category in a triplet format: \emph{<subject, action verb, object>}. Additionally, we provide segmentation masks related to hands and interaction objects.

\noindent \textbf{2. How many instances are there in total (of each type, if appropriate)?}

\textbf{A2:} The FHA-Kitchens contains 30,047 frames from 2,377 video clips, with each frame annotated for three hand sub-interaction regions, resulting in a total of 877 fine-grained action triplets and 3 coarse-grained action categories. Among them, there are 597 frames where no hand interaction action occurs, represented as {L-O\_triplet:<none>, R-O\_triplet:<none>, O-O\_triplet: <none>}.

\noindent \textbf{3. Does the dataset contain all possible instances or is it a sample (not necessarily random) of instances from a larger set? If the dataset is a sample, then what is the larger set? Is the sample representative of the larger set (e.g., geographic coverage)? If so, please describe how this representativeness was validated/verified. If it is not representative of the larger set, please describe why not (\textit{e.g.}, to cover a more diverse range of instances, because instances were withheld or unavailable).}

\textbf{A3:} FHA-Kitchens is a real-world sample of human hands part in the kitchen scenes, including information about their hand actions. The data is sourced from an existing large-scale whole-body action dataset~\cite{700_2020}, from which we selected videos featuring hand interaction actions. We extracted a total of 2,377 video clips, amounting to 84.22 minutes of footage, encompassing 8 distinct types of dishes. Due to the diversity of real-world human hand actions, it's impossible to cover all types of actions. The FHA-Kitchens dataset focuses primarily on multi-granularity hand action tasks. To address the granularity issue, we improved the hand action information in the existing dataset. Compared to the data's original annotations in Kinetics-700\_2020~\cite{700_2020}, our dataset expanded the action labels by 7 dimensions, increased the number of action categories by 52 times, and introduced 122 new action verbs. We provide a finer-grained set of hand-action instances than ever before, facilitating further research in hand-action.

\noindent \textbf{4. What data does each instance consist of? ``Raw'' data (\textit{e.g.}, unprocessed text or images)or features? In either case, please provide a description.}

\textbf{A4:} Each video frame consists of at most 9 types of bonding boxes (\textit{i.e.}, three hand sub-interaction regions and interaction objects within interaction region) and sub-interaction region corresponding coarse- and fine-grained descriptions (\textit{i.e.}, \emph{L-O}, \emph{R-O}, \emph{O-O}, and \emph{<subject, action verb, object>}). Additionally, we took into account the ``active-passive'' relationships between object pairs and the specific contact areas involved in the interaction actions. Consequently, our annotation process encompassed a total of nine dimensions, resulting in a total of 877 fine-grained hand action triplets and 3 coarse-grained hand action categories. The annotated visualizations are shown in Figure \ref{annvis} and corresponding details are listed in Figure \ref{annlist}.

\noindent \textbf{5. Is there a label or target associated with each instance? If so, please provide a description.}

\textbf{A5:} Yes. Due to our parallel annotation process, we generated annotation files in different styles. However, we consolidated all the bounding box and triplet annotation information into a single CSV file. In the merged CSV file, each instance is annotated with labels following the style of the Kinetics~\cite{6kinetic400,7kinetics600,8kinetic700,700_2020} and AVA~\cite{10ava} datasets, which include video\_name, video\_id, clip\_id, clip\_name, frame\_name, timestamp, L-O\_triplet, L-O\_action\_verb\_id, L-O\_ action\_verb\_class, L-O\_action\_bbox, left\_hand\_bbox, O1\_class, O1 \_bbox, R-O\_triplet, R-O\_action\_verb\_id, R-O\_action\_verb\_class, R-O\_action \_bbox, right \_hand\_bbox, O2\_class, O2\_bbox, O-O\_triplet, O-O\_action \_verb\_id, O-O\_action \_verb\_class, O-O\_action\_bbox, subject\_class, subject\_bbox, object\_class, object\_bbox, action\_verb \_triplet, action\_verb\_triplet\_id.


\noindent \textbf{6. Is any information missing from individual instances? If so, please provide a description, explaining why this information is missing (\textit{e.g.}, because it was unavailable). This does not include intentionally removed information but might include, \textit{e.g.}, redacted text.}

\textbf{A6:} Yes. Some instances may not have all 9 types of bonding boxes and their corresponding coarse-fine-grained action categories and segmentation annotation because of interaction action scenes, severe occlusion, truncation, blur, or small scale. We just annotated ``None'' in our annotation file to represent this situation.

\noindent \textbf{7. Are relationships between individual instances made explicit (\textit{e.g.}, users’ movie ratings, and social network links)? If so, please describe how these relationships are made explicit.}

\textbf{A7:} Yes. We provide different styles of annotation files, in COCO-style, the annotations are connected by image id and category id, you can easily access them by \href{https://github.com/jin-s13/xtcocoapi} {COCO APIs}. In CSV style, one line represents the annotations of one frame and can be processed by the pandas library easily.

\noindent \textbf{8. Are there recommended data splits (\textit{e.g.}, training, development/validation, testing)? If so, please provide a description of these splits, explaining the rationale behind them.}

\textbf{A8:} Yes. We randomly split the dataset into the disjoint train, validation, and test sets following the ratio of 7:1:2. 

\noindent \textbf{9. Are there any errors, sources of noise, or redundancies in the dataset? If so, please provide a description.}

\textbf{A9:} Although we conducted three rounds of cross-checking and corrections, there may still be some errors in the annotations, \textit{e.g.}, inappropriate bounding box annotations, or small drifts of the bounding box locations, incorrectly written verbs or nouns, insufficient granularity in verb or noun descriptions, inappropriate formatting of triplets, etc. However, we have made every effort to minimize such occurrences.

To analyze the quality of annotations, we randomly selected 500 frames and conducted manual evaluations for correctness. The results are reported in Table~\ref{error_rate_table}. These error rates are comparable to recently published datasets~\cite{1epic}.

\begin{table}[htbp]
  \centering
   \caption{Error rate in FHA-Kitchens. I-O: Interaction Objects, I-R: Interaction Regions.}
   \setlength{\tabcolsep}{1mm}{
    \begin{tabular}{cccccc}
     \toprule
    &\multicolumn{1}{l}{\textbf{Frames}} & \multicolumn{1}{l}{\textbf{I-O Boxes}} & \multicolumn{1}{l}{\textbf{I-R Boxes}} & \multicolumn{1}{l}{\textbf{Verb}} & \multicolumn{1}{l}{\textbf{Noun}} \\
    \midrule
    \textbf{Total Number} & 500	& 3,006 & 1,503 & 1,503 & 2,006 \\
    \textbf{Error Rate (\%)} & - & 4.9 & 2.5 & 2.2 & 5.3 \\
     \bottomrule
    \end{tabular}}%
  \label{error_rate_table}%
\end{table}%

\noindent \textbf{10. Is the dataset self-contained, or does it link to or otherwise rely on external resources (\textit{e.g.}, websites, tweets, other datasets)? If it links to or relies on external resources, a) are there guarantees that they will exist, and remain constant, over time; b) are there official archival versions of the complete dataset (\textit{i.g.}, including the external resources as they existed at the time the dataset was created); c) are there any restrictions (\textit{e.g.}, licenses, fees) associated with any of the external resources that might apply to a future user? Please provide descriptions of all external resources and any restrictions associated with them, as well as links or other access points, as appropriate.}

\textbf{A10:} Our dataset was derived from a large-scale publicly available dataset, namely \href{https://www.deepmind.com/open-source/kinetics} {Kinetics-700\_2020}~\cite{700_2020}, which is publicly available for download from their website. The Kinetics dataset follows the Creative Commons Attribution 4.0 International License. We would like to express our gratitude to the authors for their significant contributions to the research community.


\noindent \textbf{11. Does the dataset contain data that might be considered confidential (\textit{e.g.}, data that is protected by legal privilege or by doctor-patient confidentiality, data that includes the content of individuals' non-public communications)? If so, please provide a description.}

\textbf{A11:} No.

\noindent \textbf{12. Does the dataset contain data that, if viewed directly, might be offensive, insulting, threatening, or might otherwise cause anxiety? If so, please describe why.}

\textbf{A12:} No.

\subsubsection{Collection Process}
\ 

\noindent \textbf{1. How was the data associated with each instance acquired? Was the data directly observable (e.g., raw text, movie ratings), reported by subjects (e.g., survey responses), or indirectly inferred/derived from other data (e.g., part-of-speech tags, model-based guesses for age or language)? If data was reported by subjects or indirectly inferred/derived from other data, was the data validated/verified? If so, please describe how.}

\textbf{A1:} Our data was obtained from the existing large-scale publicly available dataset, namely \href{https://www.deepmind.com/open-source/kinetics} {Kinetics-700\_2020}~\cite{700_2020}, which is publicly available for download from their website, and then further cleaned, frame segmented, and reorganized to obtain 2377 video clips. The annotation of fine-grained action triplets was carried out on the \href{https://www.mturk.com/}{Amazon Mechanical Turk} platform, while the bounding box and coarse-grained actions annotation was facilitated using the \href{https://github.com/open-mmlab/labelbee-client}{LabelBee} tool.


\noindent \textbf{2. What mechanisms or procedures were used to collect the data (e.g., hardware apparatus or sensor, manual human curation, software program, software API)? How were these mechanisms or procedures validated?}

\textbf{A2:} The data in FHA-Kitchens come from dataset publicly available datasets described above, which can be directly downloaded from \href{https://www.deepmind.com/open-source/kinetics} {their websites}.

\noindent \textbf{3. If the dataset is a sample from a larger set, what was the sampling strategy (e.g., deterministic, probabilistic with specific sampling probabilities)?}

\textbf{A3:} Currently, we focus exclusively on hand interaction actions in kitchen scenes, thus primarily extracting data that includes hand interaction actions in kitchen scenes.

\noindent \textbf{4. Who was involved in the data collection process (\textit{e.g.}, students, crowdworkers, contractors), and how were they compensated (e.g., how much were crowdworkers paid)?}

\textbf{A4:} The authors and some volunteer undergraduate students from Wuhan University collected this dataset. The annotation compensation is based on the prevailing market rates.

\noindent \textbf{5. Over what timeframe was the data collected? Does this timeframe match the creation timeframe of the data associated with the instances (e.g., the recent crawl of old news articles)? If not, please describe the timeframe in which the data associated with the instances was created.}

\textbf{A5}: It took about 1 week to collect the data and about 6 weeks to complete organization and annotation (starting March 2023), as each participant labeled the bonding boxes and action triplets about four hours per workday. And the segmentation masks are generated by the Segment-Anything Model~\cite{20segment} guided by the bonding boxes, and corrected by human annotators for about one week.

\subsubsection{Preprocessing/cleaning/labeling}
\ 

\noindent \textbf{1. Was any preprocessing/cleaning/labeling of the data done (e.g., discretization or bucketing, tokenization, part-of-speech tagging, SIFT feature extraction, removal of instances, processing of missing values)? If so, please provide a description. If not, you may skip the remainder of the questions in this section.} 

\textbf{A1:} Yes. Since we focus on hand actions, we performed filtering and processing operations on the original videos, including the following three steps. (1) First, we observed that kitchen scenes often featured hand actions, with video content prominently showcasing human hand parts. Therefore, we sought out and extracted relevant videos that were set against a kitchen backdrop. (2) Then, to ensure the quality of the dataset, we selectively chose videos with higher resolutions. Specifically, 87\% of the videos were recorded at 1,280 $\times$ 720 resolution, while another 13\% had a shorter side of 480. Additionally, 67\% of the videos were captured at 30 frames per second (fps), and another 33\% were recorded at 24$\sim$25 fps. (3) Subsequently, we imposed a duration constraint on the videos, ranging from 30 seconds to 5 minutes, to exclude excessively long-duration videos. This constraint aimed to maintain a balanced distribution within the sample space. Finally, we collected a total of 2,377 video clips, amounting to 84.22 minutes of footage, encompassing 8 distinct types of dishes.

The collected video data was reorganized and cleaned to align with our annotation criteria. First, we split the collected video data into individual frames, as our annotated units are frames. Subsequently, we conducted further cleaning of the frames by excluding those that did not depict hands or exhibited meaningless hand actions. This cleaning process took into consideration factors such as occlusion, frame quality (\textit{i.e.}, without significant blur, subtitles, and logos), meaningful hand actions, and frame continuity. As a result, we obtained a total of 30,047 high-quality candidate video frames containing diverse hand actions for our FHA-Kitchens dataset. Compared to the initial collection, 113,436 frames were discarded during the cleaning process.

We recruited 10 voluntary annotators, whose responsibility was to annotate bounding boxes and multi-granularity action categories for each hand interaction region. To enhance annotation efficiency, we implemented a parallel annotation pipeline. The annotation of fine-grained action triplets was carried out on the \href{https://www.mturk.com/}{Amazon Mechanical Turk} platform, while the bounding box and coarse-grained actions annotation was facilitated using the \href{https://github.com/open-mmlab/labelbee-client} {LabelBee} tool. To ensure the annotation quality, three rounds of cross-checking and corrections were conducted.

\noindent \textbf{2. Was the ``raw'' data saved in addition to the preprocessed/ cleaned/ labeled data (e.g., to support unanticipated future uses)? If so, please provide a link or other access point to the ``raw'' data.}

\textbf{A2:} No.

\noindent \textbf{3. Is the software used to preprocess/clean/label the instances available? If so, please provide a link or other access point.}

\textbf{A3:} The annotation of fine-grained action triplets was carried out on the \href{https://www.mturk.com/}{Amazon Mechanical Turk} platform, while the bounding box and coarse-grained actions annotation was facilitated using the \href{https://github.com/open-mmlab/labelbee-client}{LabelBee} tool.

\subsubsection{Uses}

\ 

\noindent \textbf{1. Has the dataset been used for any tasks already? If so, please provide a description.}

\textbf{A1:} No.

\noindent \textbf{2. Is there a repository that links to any or all papers or systems that use the dataset? If so, please provide a link or other access point.}

\textbf{A2:} N/A.

\noindent \textbf{3. What (other) tasks could the dataset be used for?}

\textbf{A3:} FHA-Kitchens can be used for the research of fine-grained hand action recognition, multi-granularity hand action detection, and interaction object detection. Besides, it can also be used for specific machine learning topics such as domain generalization and action segmentation. Please see Section 3.4 of the main paper and Section ~\ref{A.1.3} of the supplementary materials.

\noindent \textbf{4. Is there anything about the composition of the dataset or the way it was collected and preprocessed/cleaned/labeled that might impact future uses? For example, is there anything that a future user might need to know to avoid uses that could result in unfair treatment of individuals or groups (e.g., stereotyping, quality of service issues) or other undesirable harms (e.g., financial harms, legal risks) If so, please provide a description. Is there anything a future user could do to mitigate these undesirable harms?}

\textbf{A4:} No.

\noindent \textbf{5. Are there tasks for which the dataset should not be used? If so, please provide a description.}

\textbf{A5:} No.

\subsubsection{Distribution}
\ 

\noindent \textbf{1. Will the dataset be distributed to third parties outside of the entity (e.g., company, institution, organization) on behalf of which the dataset was created? If so, please provide a description.}

\textbf{A1:} Yes. The dataset will be made publicly available to the research community.

\noindent \textbf{2. How will the dataset will be distributed (e.g., tarball on website, API, GitHub)? Does the dataset have a digital object identifier (DOI)?}

\textbf{A2:} It will be publicly available on the project website at \href{https://github.com/superZ678/MG-HAD}{GitHub}

\noindent \textbf{3. When will the dataset be distributed?}

\textbf{A3:} We will distribute the dataset and related code soon.

\noindent \textbf{4. Will the dataset be distributed under a copyright or other intellectual property (IP) license, and/or under applicable terms of use (ToU)? If so, please describe this license and/or ToU, and provide a link or other access point to, or otherwise reproduce, any relevant licensing terms or ToU, as well as any fees associated with these restrictions.}

\textbf{A4:} It will be distributed under the MIT license.

\noindent \textbf{5. Have any third parties imposed IP-based or other restrictions on the data associated with the instances? If so, please describe these restrictions, and provide a link or other access point to, or otherwise reproduce, any relevant licensing terms, as well as any fees associated with these restrictions.}

\textbf{A5:} No.

\noindent \textbf{6. Do any export controls or other regulatory restrictions apply to the dataset or to individual instances? If so, please describe these restrictions, and provide a link or other access point to, or otherwise reproduce, any supporting documentation.}

\textbf{A6:} No.

\subsubsection{Maintenance}
\ 

\noindent \textbf{1. Who will be supporting/hosting/ maintaining the dataset?}

\textbf{A1:} The authors.

\noindent \textbf{2. How can the owner/curator/manager of the dataset be contacted (e.g., email address)?}

\textbf{A2:} They can be contacted via email available on our dataset \href{https://github.com/superZ678/MG-HAD} {project website}.

\noindent \textbf{3. Is there an erratum? If so, please provide a link or other access point.}

\textbf{A3:} No. 

\noindent \textbf{4. Will the dataset be updated (e.g., to correct labeling errors, add new instances, delete instances)? If so, please describe how often, by whom, and how updates will be communicated to users (e.g., mailing list, GitHub)?}

\textbf{A4:} No. We have carefully three rounds of cross-checking the annotations to reduce the labeling errors. There may be very few labeling errors, which can be treated as noise.

\noindent \textbf{5. Will older versions of the dataset continue to be supported/ hosted/maintained? If so, please describe how. If not, please describe how its obsolescence will be communicated to users.}

\textbf{A5:} N/A.

\noindent \textbf{6. If others want to extend/augment/build on/contribute to the dataset, is there a mechanism for them to do so? If so, please provide a description. Will these contributions be validated/verified? If so, please describe how. If not, why not? Is there a process for communicating/distributing these contributions to other users? If so, please provide a description.}

\textbf{A6:} N/A. 

\begin{table*}[htbp]
  \centering
  \caption{Hyper-parameters used in MG-HAD.}
  \renewcommand\arraystretch{1.2}
    \begin{tabular}{c|c}
    \toprule
    Item & Value \\
    \midrule
    lr & 0.0001 \\
    lr\_backbone & 1e-05 \\
    weight\_decay & 0.0001 \\
    clip\_max\_norm & 0.1 \\
    pe\_temperature & 20 \\
    enc\_layers & 6 \\
    dec \_layers & 6 \\
    dim \_feedforward & 2048 \\
    hidden\_dim & 256 \\
    dropout & 0.0 \\
    nheads & 8 \\
    num\_queries & 900 \\
    enc\_n\_points & 4 \\
    dec\_n\_points & 4 \\
    transformer\_activation & ``relu'' \\
    set\_cost\_class & 2.0 \\
    set\_cost\_bbox & 5.0 \\
    set\_cost\_giou & 2.0 \\
    cls\_loss\_coef & 1.0 \\
    bbox\_loss\_coef & 5.0 \\
    giou\_loss\_coef & 2.0 \\
    focal\_alpha & 0.25 \\
    dn\_box\_noise\_scale ($\gamma_1$) & 1.0 \\
    dn\_label\_noise\_scale ($\gamma_2$) & 0.5 \\
    subject\_weight ($w_s$) & 0.2 \\
    action\_weight ($w_a$) & 0.6 \\
    object\_weight ($w_o$) & 0.2 \\    
    \bottomrule
   \end{tabular}
   \label{tab:hp}
\end{table*}

\begin{table*}[ht!]
  \centering
  \caption{Vocabulary of fine-grained hand action verbs.}
  \label{verbid}
    \footnotesize
    \centering
    \resizebox{0.5\linewidth}{!}{\begin{tabular}{lll|lll}
    \toprule
    ID & Verb & \#Instance & ID & Verb & \#Instance \\
    \hline
    0 &	hold-around	& 1,593 & 65	& knead	& 298 \\
    1 & hold-at & 788 & 66	& contain & 144\\
    2 & fill-with &	265	 & 67	& roll-on & 82 \\
    3 & pinch-on & 1,115	& 68 & stick-to	& 50 \\
    4 &	rub-around & 45	& 69 & touch-to	& 12 \\
    5 & hold-in	& 20,520& 70 & smooth-out & 144 \\
    6 & touch-on & 42	& 71 & sprinkle-on & 203 \\
    7 & hold-with & 341 & 72 & squeeze-around & 189 \\
    8 & press-on & 9,369 & 73 & press-down & 675 \\
    9 & cut-out	& 889 	& 74 & cut-up & 100 \\
    10	& fix-on & 2,037 & 75	& shovel-up	& 123 \\
    11	& peel-off	& 1,413	& 76 & grab-out	& 1 \\
    12	& slice-along & 1,306	& 77 & close & 51 \\
    13	& grab & 2,531 & 78 & rotate & 51 \\
    14	& cut-half & 230 &79& open & 72 \\
    15	& take-up & 134 & 80 & open-down	& 27 \\
    16	& pinch	& 609 & 81 & hold-down	& 41 \\
    17	& catch	& 446 & 82 & cut-dice	& 443\\
    18	& put-down & 1,406& 83 & dig-seeds & 124  \\
    19	& roll-up & 1,296 & 84 & chop & 346 \\
    20	& fix & 293 & 85 & push-forward & 8 \\
    21  & scrub-inside & 53 & 86& cut-halves & 22 \\
    22	& lay-down & 111& 87 & peel & 87 \\
    23	& hold-onto	& 453& 88 & push-ahead	& 2 \\
    24	& pick-up & 526	& 89 & screw-on	& 5 \\
    25	& cut-slice	& 2,808 & 90 & sprinkle-into & 16 \\
    26	& take-out & 178 & 91 & scoop-up & 85 \\
    27	& turn-off & 22 & 92 & hold-along & 129 \\
    28	& cut-down & 1,040 	& 93 &  scrape-on & 331 \\
    29	& cut-off & 820& 94 & stick-with & 13 \\
    30	& grab-up & 115	& 95 & cut-in & 275 \\
    31	& put-up & 56 & 96 & rub-on & 11 \\
    32	& break-apart & 221	& 97 & put-on & 2\\
    33	& touch	& 483	& 98 & push-off & 10  \\
    34	& cut-into-halves & 136 &99 & place-on & 17 \\
    35	& bring-up & 140 & 100 & cut & 15 \\
    36	& pour-out & 832 & 101 & dip-in & 9 \\
    37	& pour-into	& 395 & 102 & stretch-out & 31 \\
    38	& pour & 154 & 103	& flip & 8 \\
    39	& scrape & 70 & 104	 & set-aside	& 22  \\
    40	& rotate-around	& 162 &105& julienne & 165 \\
    41	& screw-down & 19 & 106	& unroll & 270 \\
    42	& remove-out & 69 & 107	& adjust & 46 \\
    43	& hold-up & 664	 & 108	& place-down & 262 \\
    44	& scoop-out	& 76& 109 & pile & 49 \\
    45	& open-up & 21 & 110 & pull & 137 \\
    46	& hold & 946& 111 & attach-to & 18 \\
    47	& hold-on & 187 & 112 & grab-in	& 15 \\
    48	& squeeze & 334	& 113 & knock-on & 29  \\
    49	& squeeze-out & 254& 114 & press-against & 13 \\
    50	& mix-together & 187& 115 & stir-in & 114 \\
    51	& spread-on	& 342 & 116	& pull-up & 47  \\
    52	& twist-off	& 20  & 117	& point-at & 25 \\
    53	& wrap-around & 204 & 118	& pull-out & 33 \\
    54	& break-off	& 417 & 119 & scrape-down & 6 \\
    55	& grab-at & 621 & 120	& grab-onto & 558 \\
    56	& grab-on & 384	& 121 & hold-into & 88 \\
    57	& cut-through & 25	& 122 & hold-over & 67 \\
    58  & chop-dice	& 63 & 123 & stir-into & 49 \\
    58  & chop-dice	& 63 & 124 & press-onto	& 1 \\
    60	& insert-into & 106 & 125 & roll & 68 \\
    61	& put-in & 30  & 126 & roll-out & 33 \\
    62	& dig-out & 23& 127 & dip & 51 \\
    63	& cut-chunks & 91& 128 & brush-onto & 54 \\
    64	& churn	& 369 & 129	& flip-over & 83 \\	
    \bottomrule
   \end{tabular}}
\end{table*}

\begin{table*}[htbp]
  \centering
  \caption{Vocabulary of fine-grained interaction object nouns.}
  \label{nounid1}
    \centering
    \resizebox{0.8\linewidth}{!}{
    \begin{tabular}{c|lll|c|lll}
    \toprule
     Super category & ID & Noun & \#Instance & Super category & ID & Noun & \#Instance \\
    \hline
    \multirow{65}{*}{Vegetables\&Plants}	&	0	&	basil-end	&	201	&	\multirow{47}{*}{Fruits}	&	65	&	apple-all	&	22	\\
    \multirow{65}{*}{~}	&	1	&	beet-end	&	143	&	\multirow{47}{*}{~}	&	66	&	apple-end	&	34	\\
    \multirow{65}{*}{~}	&	2	&	beet-head	&	136	&	\multirow{47}{*}{~}	&	67	&	apple-head	&	253	\\
    \multirow{65}{*}{~}	&	3	&	beet-middle	&	36	&	\multirow{47}{*}{~}	&	68	&	apple-middle	&	301	\\
    \multirow{65}{*}{~}	&	4	&	bell-pepper-all	&	2	&	\multirow{47}{*}{~}	&	69	&	avocado-end	&	156	\\
    \multirow{65}{*}{~}	&	5	&	bell-pepper-end	&	210	&	\multirow{47}{*}{~}	&	70	&	avocado-head	&	12	\\
    \multirow{65}{*}{~}	&	6	&	bell-pepper-head	&	67	&	\multirow{47}{*}{~}	&	71	&	avocado-left	&	34	\\
    \multirow{65}{*}{~}	&	7	&	bell-pepper-middle	&	139	&	\multirow{47}{*}{~}	&	72	&	avocado-middle	&	174	\\
    \multirow{65}{*}{~}	&	8	&	broccoli-head	&	38	&	\multirow{47}{*}{~}	&	73	&	avocado-right	&	34	\\
    \multirow{65}{*}{~}	&	9	&	carrot-end	&	1,625	&	\multirow{47}{*}{~}	&	74	&	block-watermelon-edge	&	5	\\
    \multirow{65}{*}{~}	&	10	&	carrot-head	&	205	&	\multirow{47}{*}{~}	&	75	&	green-melon-all	&	4,074	\\
    \multirow{65}{*}{~}	&	11	&	carrot-middle	&	638	&	\multirow{47}{*}{~}	&	76	&	green-melon-end	&	68	\\
    \multirow{65}{*}{~}	&	12	&	chopped-vegetables-surface	&	38	&	\multirow{47}{*}{~}	&	77	&	green-melon-middle	&	68	\\
    \multirow{65}{*}{~}	&	13	&	courgette-end	&	921	&	\multirow{47}{*}{~}	&	78	&	half-apple-head	&	139	\\
    \multirow{65}{*}{~}	&	14	&	courgette-middle	&	39	&	\multirow{47}{*}{~}	&	79	&	half-apple-middle	&	47	\\
    \multirow{65}{*}{~}	&	15	&	cucumber-end	&	288	&	\multirow{47}{*}{~}	&	80	&	half-pineapple-head	&	6	\\
    \multirow{65}{*}{~}	&	16	&	cucumber-middle	&	165	&	\multirow{47}{*}{~}	&	81	&	half-pineapple-middle	&	99	\\
    \multirow{65}{*}{~}	&	17	&	cucumber-strip-all	&	14	&	\multirow{47}{*}{~}	&	82	&	half-tomato-end	&	1	\\
    \multirow{65}{*}{~}	&	18	&	cucumber-strip-end	&	72	&	\multirow{47}{*}{~}	&	83	&	half-tomato-middle	&	1	\\
    \multirow{65}{*}{~}	&	19	&	cucumber-strip-middle	&	22	&	\multirow{47}{*}{~}	&	84	&	half-watermelon-edge	&	79	\\
    \multirow{65}{*}{~}	&	20	&	garlic-middle	&	240	&	\multirow{47}{*}{~}	&	85	&	half-watermelon-end	&	17	\\
    \multirow{65}{*}{~}	&	21	&	garlic-end	&	164	&	\multirow{47}{*}{~}	&	86	&	half-watermelon-head	&	301	\\
    \multirow{65}{*}{~}	&	22	&	garlic-head	&	46	&	\multirow{47}{*}{~}	&	87	&	half-watermelon-middle	&	232	\\
    \multirow{65}{*}{~}	&	23	&	ginger-end	&	248	&	\multirow{47}{*}{~}	&	88	&	lemon-end	&	156	\\
    \multirow{65}{*}{~}	&	24	&	ginger-head	&	169	&	\multirow{47}{*}{~}	&	89	&	lemon-middle	&	108	\\
    \multirow{65}{*}{~}	&	25	&	ginger-middle	&	70	&	\multirow{47}{*}{~}	&	90	&	melon-skin-all	&	119	\\
    \multirow{65}{*}{~}	&	26	&	green-beans-end	&	937	&	\multirow{47}{*}{~}	&	91	&	melon-skin-end	&	576	\\
    \multirow{65}{*}{~}	&	27	&	green-pepper-dice	&	1	&	\multirow{47}{*}{~}	&	92	&	melon-pulp-all	&	28	\\
    \multirow{65}{*}{~}	&	28	&	green-pepper-end	&	710	&	\multirow{47}{*}{~}	&	93	&	melon-pulp-end	&	163	\\
    \multirow{65}{*}{~}	&	29	&	green-pepper-head	&	142	&	\multirow{47}{*}{~}	&	94	&	melon-pulp-middle	&	49	\\
    \multirow{65}{*}{~}	&	30	&	green-pepper-middle	&	505	&	\multirow{47}{*}{~}	&	95	&	melon-slice-end	&	200	\\
    \multirow{65}{*}{~}	&	31	&	half-bell-pepper-end	&	116	&	\multirow{47}{*}{~}	&	96	&	orange-all	&	22	\\
    \multirow{65}{*}{~}	&	32	&	half-bell-pepper-middle	&	110	&	\multirow{47}{*}{~}	&	97	&	orange-end	&	24	\\
    \multirow{65}{*}{~}	&	33	&	half-onion-all	&	23	&	\multirow{47}{*}{~}	&	98	&	orange-head	&	209	\\
    \multirow{65}{*}{~}	&	34	&	half-onion-head	&	11	&	\multirow{47}{*}{~}	&	99	&	orange-middle	&	547	\\
    \multirow{65}{*}{~}	&	35	&	half-onion-middle	&	11	&	\multirow{47}{*}{~}	&	100	&	peelless-orange-middle	&	93	\\
    \multirow{65}{*}{~}	&	36	&	mushroom-middle	&	15	&	\multirow{47}{*}{~}	&	101	&	piece-orange-edge	&	276	\\
    \multirow{65}{*}{~}	&	37	&	nori-all	&	506	&	\multirow{47}{*}{~}	&	102	&	pineapple-all	&	26	\\
    \multirow{65}{*}{~}	&	38	&	nori-end	&	262	&	\multirow{47}{*}{~}	&	103	&	pineapple-end	&	476	\\
    \multirow{65}{*}{~}	&	39	&	onion-end	&	78	&	\multirow{47}{*}{~}	&	104	&	pineapple-head	&	959	\\
    \multirow{65}{*}{~}	&	40	&	onion-head	&	28	&	\multirow{47}{*}{~}	&	105	&	pineapple-middle	&	1,346	\\
    \multirow{65}{*}{~}	&	41	&	onion-middle	&	49	&	\multirow{47}{*}{~}	&	106	&	slice-pineapple-end	&	25	\\
    \multirow{65}{*}{~}	&	42	&	pepper-seeds-all	&	4	&	\multirow{47}{*}{~}	&	107	&	slice-pineapple-middle	&	63	\\
    \multirow{65}{*}{~}	&	43	&	piece-onion-middle	&	38	&	\multirow{47}{*}{~}	&	108	&	watermelon-edge	&	29	\\
    \multirow{65}{*}{~}	&	44	&	piece-tomato-end	&	41	&	\multirow{47}{*}{~}	&	109	&	watermelon-end	&	966	\\
    \multirow{65}{*}{~}	&	45	&	purple-cabbage-end	&	77	&	\multirow{47}{*}{~}	&	110	&	watermelon-head	&	155	\\
    \multirow{65}{*}{~}	&	46	&	purple-cabbage-head	&	70	&	\multirow{47}{*}{~}	&	111	&	watermelon-middle	&	631	\\
    \cline{5-8}
    \multirow{65}{*}{~}	&	47	&	purple-cabbage-middle	&	23	&	\multirow{12}{*}{Dairy\&Eggs}	&	112	&	boiled-egg-end	&	226	\\
    \multirow{65}{*}{~}	&	48	&	red-pepper-all	&	21	&	\multirow{12}{*}{~}	&	113	&	boiled-egg-head	&	22	\\
    \multirow{65}{*}{~}	&	49	&	red-pepper-head	&	25	&	\multirow{12}{*}{~}	&	114	&	boiled-egg-middle	&	14	\\
    \multirow{65}{*}{~}	&	50	&	red-pepper-middle	&	18	&	\multirow{12}{*}{~}	&	115	&	boiled-egg-shell	&	88	\\
    \multirow{65}{*}{~}	&	51	&	small-tomato-head	&	9	&	\multirow{12}{*}{~}	&	116	&	egg-all	&	248	\\
    \multirow{65}{*}{~}	&	52	&	small-tomato-middle	&	27	&	\multirow{12}{*}{~}	&	117	&	egg-head	&	1	\\
    \multirow{65}{*}{~}	&	53	&	spinach-end	&	4	&	\multirow{12}{*}{~}	&	118	&	egg-middle	&	26	\\
    \multirow{65}{*}{~}	&	54	&	spinach-head	&	35	&	\multirow{12}{*}{~}	&	119	&	egg-liquid-all	&	10	\\
    \multirow{65}{*}{~}	&	55	&	spinach-middle	&	30	&	\multirow{12}{*}{~}	&	120	&	egg-shell-all	&	86	\\
    \multirow{65}{*}{~}	&	56	&	spring-garlic-all	&	18	&	\multirow{12}{*}{~}	&	121	&	egg-shell-edge	&	34	\\
    \multirow{65}{*}{~}	&	57	&	spring-garlic-end	&	53	&	\multirow{12}{*}{~}	&	122	&	milk-all	&	594	\\
    \multirow{65}{*}{~}	&	58	&	spring-garlic-head	&	23	&	\multirow{12}{*}{~}	&	123	&	yolk-all	&	112	\\
    \cline{5-8}
    \multirow{65}{*}{~}	&	59	&	spring-garlic-middle	&	52	& \multirow{6}{*}{Meat\&Fish}	&	124	&	chicken-dice	&	8 \\
    \multirow{65}{*}{~}	&	60	&	sun-flower-seeds	&	104	&	\multirow{6}{*}{~}	&	125	&	raw-chicken-dice	&	69 \\
    \multirow{65}{*}{~}	&	61	&	tomato-cube	&	22	&	\multirow{6}{*}{~}	&	126	&	crab-shred	&	328 \\
    \multirow{65}{*}{~}	&	62	&	tomato-end	&	280	&	\multirow{6}{*}{~}	&	127	&	meat-end	&	515 \\
    \multirow{65}{*}{~}	&	63	&	tomato-middle	&	17	&	\multirow{6}{*}{~}	&	128	&	meat-head	&	573 \\
    \multirow{65}{*}{~}	&	64	&	tomato-sliced-middle	&	109	&	\multirow{6}{*}{~}	&	129	&	meat-middle	&	142 \\
    \hline
   \end{tabular}}
\end{table*}

\begin{table*}[htbp]
  \centering
  \caption{Vocabulary of fine-grained interaction object nouns.}
  \label{nounid2}
    \centering
    \resizebox{0.8\linewidth}{!}{ \begin{tabular}{c|lll|c|lll}
    \toprule
     Super category & ID & Noun & \#Instance & Super category & ID & Noun & \#Instance \\
    \hline
    \multirow{9}{*}{Meat\&Fish}	&	130	&	meat-piece-end	&	442	&	\multirow{49}{*}{Containers}	&	194	&	bottle-body	&	52	\\
    \multirow{9}{*}{~}	&	131	&	meat-slice-all	&	1	&	\multirow{49}{*}{~}	&	195	&	box-lid-bottom	&	21	\\
    \multirow{9}{*}{~}	&	132	&	meat-slice-end	&	202	&	\multirow{49}{*}{~}	&	196	&	bottle-cap	&	8	\\
    \multirow{9}{*}{~}	&	133	&	piece-pepperoni-all	&	87	&	\multirow{49}{*}{~}	&	197	&	bottle-cap-all	&	106	\\
    \multirow{9}{*}{~}	&	134	&	piece-pepperoni-end	&	158	&	\multirow{49}{*}{~}	&	198	&	bottle-cap-bottom	&	19	\\
    \multirow{9}{*}{~}	&	135	&	salmon-piece-all	&	36	&	\multirow{49}{*}{~}	&	199	&	can-cover-edg	&	2	\\
    \multirow{9}{*}{~}	&	136	&	salmon-piece-end	&	399	&	\multirow{49}{*}{~}	&	200	&	can-cover-edge	&	106	\\
    \multirow{9}{*}{~}	&	137	&	salmon-piece-middle	&	14	&	\multirow{49}{*}{~}	&	201	&	ceramic-cup-all	&	68	\\
    \multirow{9}{*}{~}	&	138	&	salmon-slice-end	&	44	&	\multirow{49}{*}{~}	&	202	&	ceramic-cup-body	&	671	\\
    \cline{1-4}
    \multirow{13}{*}{Spices\&Sauces}	&	139	&	butter-all	&	45	&	\multirow{49}{*}{~}	&	203	&	ceramic-cup-handle	&	12	\\
    \multirow{13}{*}{~}	&	140	&	crumbles-cheese-all	&	116	&	\multirow{49}{*}{~}	&	204	&	ceramic-lid-all	&	25	\\
    \multirow{13}{*}{~}	&	141	&	cheese-all	&	27	&	\multirow{49}{*}{~}	&	205	&	ceramic-lid-edge	&	85	\\
    \multirow{13}{*}{~}	&	142	&	green-butter-all	&	85	&	\multirow{49}{*}{~}	&	206	&	ceramic-teapot-handle	&	132	\\
    \multirow{13}{*}{~}	&	143	&	mozzarella-all	&	84	&	\multirow{49}{*}{~}	&	207	&	ceramic-teacup-body	&	69	\\
    \multirow{13}{*}{~}	&	144	&	mozzarella-end	&	12	&	\multirow{49}{*}{~}	&	208	&	ceramic-teacup-edge	&	138	\\
    \multirow{13}{*}{~}	&	145	&	pizza-sauce-end	&	193	&	\multirow{49}{*}{~}	&	209	&	cup-edge	&	79	\\
    \multirow{13}{*}{~}	&	146	&	powder-all	&	29	&	\multirow{49}{*}{~}	&	210	&	glass-bottle-edge	&	40	\\
    \multirow{13}{*}{~}	&	147	&	sauce-all	&	397	&	\multirow{49}{*}{~}	&	211	&	glass-bottle-top	&	59	\\
    \multirow{13}{*}{~}	&	148	&	slice-cheese-end	&	44	&	\multirow{49}{*}{~}	&	212	&	glass-cup-body	&	1	\\
    \multirow{13}{*}{~}	&	149	&	tomato-sauce-all	&	151	&	\multirow{49}{*}{~}	&	213	&	glass-cup-edge	&	217	\\
    \multirow{13}{*}{~}	&	150	&	tomato-sauce-edge	&	15	&	\multirow{49}{*}{~}	&	214	&	glass-cup-handle	&	51	\\
    \multirow{13}{*}{~}	&	151	&	sauce-mixed	&	70	&	\multirow{49}{*}{~}	&	215	&	glass-goblet-stem	&	333	\\
    \cline{1-4}
    \multirow{5}{*}{Liquids}	&	152	&	can-opener	&	108	&	\multirow{49}{*}{~}	&	216	&	glastic-bottle-edge	&	4	\\
    \multirow{5}{*}{~}	&	153	&	green-mixture-all	&	248	&	\multirow{49}{*}{~}	&	217	&	glastic-bottle-top	&	4	\\
    \multirow{5}{*}{~}	&	154	&	jam-all	&	23	&	\multirow{49}{*}{~}	&	218	&	grass-bottle-top	&	30	\\
    \multirow{5}{*}{~}	&	155	&	oil-all	&	60	&	\multirow{49}{*}{~}	&	219	&	iron-basin-body	&	29	\\
    \multirow{5}{*}{~}	&	156	&	olive-oil-all	&	51	&	\multirow{49}{*}{~}	&	220	&	iron-basin-edge	&	145	\\
    \cline{1-4}
    \multirow{3}{*}{Baked\&Baking}	&	157	&	baking-paper-edge	&	99	&	\multirow{49}{*}{~}	&	221	&	iron-basin-middle	&	115	\\
    \multirow{3}{*}{~}	&	158	&	baking-paper-top	&	25	&	\multirow{49}{*}{~}	&	222	&	iron-cup-body	&	1,005	\\
    \multirow{3}{*}{~}	&	159	&	baking-plate-edge	&	54	&	\multirow{49}{*}{~}	&	223	&	iron-cup-handle	&	325	\\
    \cline{1-4}
    \multirow{15}{*}{Cooked Food}	&	160	&	piece-pizza-end	&	125	&	\multirow{49}{*}{~}	&	224	&	iron-dipper-handle	&	14	\\
    \multirow{15}{*}{~}	&	161	&	pizza-all	&	63	&	\multirow{49}{*}{~}	&	225	&	plastic-basin-edge	&	33	\\
    \multirow{15}{*}{~}	&	162	&	pizza-end	&	27	&	\multirow{49}{*}{~}	&	226	&	plastic-bottle-bottom	&	21	\\
    \multirow{15}{*}{~}	&	163	&	pizza-middle	&	29	&	\multirow{49}{*}{~}	&	227	&	plastic-bottle-edge	&	352	\\
    \multirow{15}{*}{~}	&	164	&	sandwich-edge	&	32	&	\multirow{49}{*}{~}	&	228	&	plastic-bottle-top	&	78	\\
    \multirow{15}{*}{~}	&	165	&	sandwich-end	&	297	&	\multirow{49}{*}{~}	&	229	&	plastic-cup-body	&	19	\\
    \multirow{15}{*}{~}	&	166	&	sandwich-head	&	141	&	\multirow{49}{*}{~}	&	230	&	sauce-container-end	&	4	\\
    \multirow{15}{*}{~}	&	167	&	sandwich-middle	&	123	&	\multirow{49}{*}{~}	&	231	&	small-cup-edge	&	195	\\
    \multirow{15}{*}{~}	&	168	&	sandwich-side	&	85	&	\multirow{49}{*}{~}	&	232	&	small-plastic-bottle-edge	&	154	\\
    \multirow{15}{*}{~}	&	169	&	sandwich-top	&	27	&	\multirow{49}{*}{~}	&	233	&	small-plastic-bottle-end	&	120	\\
    \multirow{15}{*}{~}	&	170	&	sandwich-all	&	23	&	\multirow{49}{*}{~}	&	234	&	small-plastic-bottle-top	&	167	\\
    \multirow{15}{*}{~}	&	171	&	sushi-roll-all	&	151	&	\multirow{49}{*}{~}	&	235	&	teapot-lid-edge	&	126	\\
    \multirow{15}{*}{~}	&	172	&	sushi-roll-end	&	1,659	&	\multirow{49}{*}{~}	&	236	&	teapot-lid-handle	&	267	\\
    \multirow{15}{*}{~}	&	173	&	sushi-roll-head	&	233	&	\multirow{49}{*}{~}	&	237	&	wine-bottle-bottom	&	75	\\
    \multirow{15}{*}{~}	&	174	&	sushi-roll-middle	&	622	&	\multirow{49}{*}{~}	&	238	&	yogurt-box-bottom	&	63	\\
    \cline{1-4}
    \multirow{15}{*}{Packaging}	&	175	&	bamboo-mat-edge	&	284	&	\multirow{49}{*}{~}	&	239	&	yogurt-box-edge	&	62	\\
    \multirow{15}{*}{~}	&	176	&	bamboo-mat-end	&	528	&	\multirow{49}{*}{~}	&	240	&	yogurt-box-handle	&	2	\\
    \multirow{15}{*}{~}	&	177	&	bamboo-mat-head	&	8	&	\multirow{49}{*}{~}	&	241	&	yogurt-box-top	&	14	\\
    \multirow{15}{*}{~}	&	178	&	bamboo-mat-middle	&	244	&	\multirow{49}{*}{~}	&	242	&	sauce-cup-all	&	9	\\
    \cline{5-8}
    \multirow{15}{*}{~}	&	179	&	mozzarella-bag-end	&	71	&	\multirow{14}{*}{Cutlery}	&	243	&	bowl-bottom	&	69	\\
    \multirow{15}{*}{~}	&	180	&	mozzarella-bag-middle	&	24	&	\multirow{14}{*}{~}	&	244	&	bowl-edge	&	198	\\
    \multirow{15}{*}{~}	&	181	&	onion-bag-end	&	86	&	\multirow{14}{*}{~}	&	245	&	glass-bowl-all	&	23	\\
    \multirow{15}{*}{~}	&	182	&	onion-bag-middle	&	30	&	\multirow{14}{*}{~}	&	246	&	glass-bowl-body	&	23	\\
    \multirow{15}{*}{~}	&	183	&	pepperoni-bag-end	&	60	&	\multirow{14}{*}{~}	&	247	&	glass-bowl-bottom	&	91	\\
    \multirow{15}{*}{~}	&	184	&	pepperoni-bag-middle	&	16	&	\multirow{14}{*}{~}	&	248	&	glass-bowl-edge	&	415	\\
    \multirow{15}{*}{~}	&	185	&	piping-bag-all	&	251	&	\multirow{14}{*}{~}	&	249	&	glass-bowl-handle	&	8	\\
    \multirow{15}{*}{~}	&	186	&	pizza-box-edge	&	140	&	\multirow{14}{*}{~}	&	250	&	grass-bowl-edge	&	13	\\
    \multirow{15}{*}{~}	&	187	&	tea-leaves-bag-body	&	38	&	\multirow{14}{*}{~}	&	251	&	green-bowl-edge	&	29	\\
    \multirow{15}{*}{~}	&	188	&	tea-leaves-bag-bottom	&	42	&	\multirow{14}{*}{~}	&	252	&	small-bowl-edge	&	16	\\
    \multirow{15}{*}{~}	&	189	&	tea-leaves-bag-top	&	38	&	\multirow{14}{*}{~}	&	253	&	steel-bowl-edge	&	153	\\
    \cline{1-4}
    \multirow{4}{*}{Containers}	&	190	&	black-bottle-top	&	42	&	\multirow{14}{*}{~}	&	254	&	steel-bowl-top	&	13	\\
    \multirow{4}{*}{~}	&	191	&	bottle-all	&	1	&	\multirow{14}{*}{~}	&	255	&	metal-bowl-edge	&	470	\\
    \multirow{4}{*}{~}	&	192	&	bottle-edge	&	19	&	\multirow{14}{*}{~}	&	256	&	plastic-bowl-all	&	46	\\
    \multirow{4}{*}{~}	&	193	&	bottle-top	&	30	&		&		&		&		\\
    \bottomrule
   \end{tabular}}  
\end{table*}

\begin{table*}[htbp]
  \centering
  \caption{Vocabulary of fine-grained interaction object nouns.}
  \label{nounid3}
    \centering
    \resizebox{0.8\linewidth}{!}{ 
    \begin{tabular}{c|lll|c|lll}
    \toprule
    Super category & ID & Noun & \#Instance & Super category & ID & Noun & \#Instance \\
    \hline
    \multirow{24}{*}{Cutlery}	&	257	&	plastic-bowl-body	&	10	&	\multirow{16}{*}{Kitchenware}	&	321	&	shovel-body	&	49	\\
    \multirow{24}{*}{~}	&	258	&	plastic-bowl-edge	&	31	&	\multirow{16}{*}{~}	&	322	&	shovel-handle	&	49	\\
    \multirow{24}{*}{~}	&	259	&	porcelain-bowl-edge	&	39	&	\multirow{16}{*}{~}	&	323	&	sieve-spoon-body	&	13	\\
    \multirow{24}{*}{~}	&	260	&	porcelain-bowl-middle	&	2	&	\multirow{16}{*}{~}	&	324	&	sieve-spoon-handle	&	12	\\
    \multirow{24}{*}{~}	&	261	&	porcelain-bowl-top	&	17	&	\multirow{16}{*}{~}	&	325	&	small-iron-pot-handle	&	90	\\
    \multirow{24}{*}{~}	&	262	&	fork-handle	&	9	&	\multirow{16}{*}{~}	&	326	&	small-knife-body	&	663	\\
    \multirow{24}{*}{~}	&	263	&	iron-spoon-body	&	353	&	\multirow{16}{*}{~}	&	327	&	small-knife-handle	&	706	\\
    \multirow{24}{*}{~}	&	264	&	iron-spoon-handle	&	395	&	\multirow{16}{*}{~}	&	328	&	tea-strainer-body	&	108	\\
    \multirow{24}{*}{~}	&	265	&	plastic-scoop-body	&	48	&	\multirow{16}{*}{~}	&	329	&	tea-strainer-edge	&	4	\\
    \multirow{24}{*}{~}	&	266	&	plastic-scoop-handle	&	94	&	\multirow{16}{*}{~}	&	330	&	tea-strainer-handle	&	224	\\
    \multirow{24}{*}{~}	&	267	&	plastic-spoon-body	&	22	&	\multirow{16}{*}{~}	&	331	&	turnplate-corner	&	15	\\
    \multirow{24}{*}{~}	&	268	&	plastic-spoon-handle	&	22	&	\multirow{16}{*}{~}	&	332	&	turnplate-edge	&	9	\\
    \multirow{24}{*}{~}	&	269	&	spoon-body	&	692	&	\multirow{16}{*}{~}	&	333	&	utility-knife-body	&	10,419	\\
    \multirow{24}{*}{~}	&	270	&	spoon-handle	&	1,011	&	\multirow{16}{*}{~}	&	334	&	utility-knife-handle	&	11,157	\\
    \multirow{24}{*}{~}	&	271	&	tablespoon-body	&	327	&	\multirow{16}{*}{~}	&	335	&	carrot-peeler-body	&	255	\\
    \multirow{24}{*}{~}	&	272	&	tablespoon-handle	&	332	&	\multirow{16}{*}{~}	&	336	&	carrot-peeler-handle	&	518	\\
    \cline{5-8}
    \multirow{24}{*}{~}	&	273	&	teaspoon-body	&	51	&	\multirow{3}{*}{Appliances}	&	337	&	grater-body	&	70	\\
    \multirow{24}{*}{~}	&	274	&	teaspoon-handle	&	191	&	\multirow{3}{*}{~}	&	338	&	grater-handle	&	70	\\
    \multirow{24}{*}{~}	&	275	&	wooden-spoon-body	&	116	&	\multirow{3}{*}{~}	&	339	&	oven-door-handle	&	22	\\
    \cline{5-8}
    \multirow{24}{*}{~}	&	276	&	wooden-spoon-handle	&	152	&	\multirow{20}{*}{Rice\&Flour}	&	340	&	ball-dough-end	&	27	\\
    \multirow{24}{*}{~}	&	277	&	wooden-spatula-body	&	25	&	\multirow{20}{*}{~}	&	341	&	ball-dough-head	&	24	\\
    \multirow{24}{*}{~}	&	278	&	wooden-spatula-handle	&	28	&	\multirow{20}{*}{~}	&	342	&	ball-dough-middle	&	173	\\
    \multirow{24}{*}{~}	&	279	&	table-knife-handle	&	34	&	\multirow{20}{*}{~}	&	343	&	dough-all	&	413	\\
    \multirow{24}{*}{~}	&	280	&	tableware-handle	&	11	&	\multirow{20}{*}{~}	&	344	&	dough-flour-all	&	117	\\
    \cline{1-4}
    \multirow{40}{*}{Kitchenware}	&	281	&	beater-end	&	25	&	\multirow{20}{*}{~}	&	345	&	dough-flour-middle	&	35	\\
    \multirow{40}{*}{~}	&	282	&	plate-edge	&	335	&	\multirow{20}{*}{~}	&	346	&	flat-dough-all	&	42	\\
    \multirow{40}{*}{~}	&	283	&	plate-end	&	20	&	\multirow{20}{*}{~}	&	347	&	flat-dough-edge	&	206	\\
    \multirow{40}{*}{~}	&	284	&	brush-body	&	105	&	\multirow{20}{*}{~}	&	348	&	flat-dough-end	&	963	\\
    \multirow{40}{*}{~}	&	285	&	brush-handle	&	195	&	\multirow{20}{*}{~}	&	349	&	flat-dough-middle	&	13	\\
    \multirow{40}{*}{~}	&	286	&	can-opener-edge	&	83	&	\multirow{20}{*}{~}	&	350	&	flour-all	&	35	\\
    \multirow{40}{*}{~}	&	287	&	can-opener-end	&	25	&	\multirow{20}{*}{~}	&	351	&	green-dough-all	&	123	\\
    \multirow{40}{*}{~}	&	288	&	ceramic-plate-all	&	91	&	\multirow{20}{*}{~}	&	352	&	little-dough-al	&	1	\\
    \multirow{40}{*}{~}	&	289	&	ceramic-plate-edge	&	208	&	\multirow{20}{*}{~}	&	353	&	little-dough-all	&	386	\\
    \multirow{40}{*}{~}	&	290	&	chopping-board-edge	&	2	&	\multirow{20}{*}{~}	&	354	&	oval-dough-end	&	170	\\
    \multirow{40}{*}{~}	&	291	&	cooking-spoon-body	&	77	&	\multirow{20}{*}{~}	&	355	&	rice-all	&	585	\\
    \multirow{40}{*}{~}	&	292	&	cooking-spoon-handle	&	382	&	\multirow{20}{*}{~}	&	356	&	slice-bread-end	&	92	\\
    \multirow{40}{*}{~}	&	293	&	food-mixer-handle	&	372	&	\multirow{20}{*}{~}	&	357	&	bread-end	&	510	\\
    \multirow{40}{*}{~}	&	294	&	food-mixer	&	369	&	\multirow{20}{*}{~}	&	358	&	bread-head	&	199	\\
    \multirow{40}{*}{~}	&	295	&	pizza-cutter-body	&	29	&	\multirow{20}{*}{~}	&	359	&	bread-middle	&	258	\\
    \cline{5-8}
    \multirow{40}{*}{~}	&	296	&	pizza-cutter-handle	&	17	&	\multirow{17}{*}{Dessert}	&	360	&	chocolate-cake-edge	&	242	\\
    \multirow{40}{*}{~}	&	297	&	food-plate-center	&	75	&	\multirow{17}{*}{~}	&	361	&	chocolate-cake-top	&	111	\\
    \multirow{40}{*}{~}	&	298	&	food-plate-edge	&	36	&	\multirow{17}{*}{~}	&	362	&	chocolate-all	&	21	\\
    \multirow{40}{*}{~}	&	299	&	handle	&	108	&	\multirow{17}{*}{~}	&	363	&	candies-all	&	159	\\
    \multirow{40}{*}{~}	&	300	&	iron-plate-edge	&	76	&	\multirow{17}{*}{~}	&	364	&	candy-all	&	4	\\
    \multirow{40}{*}{~}	&	301	&	kitchen-knife-body	&	552	&	\multirow{17}{*}{~}	&	365	&	candy-top	&	51	\\
    \multirow{40}{*}{~}	&	302	&	kitchen-knife-handle	&	1,215	&	\multirow{17}{*}{~}	&	366	&	chocolate-bar-middle	&	87	\\
    \multirow{40}{*}{~}	&	303	&	knife-body	&	26	&	\multirow{17}{*}{~}	&	367	&	chocolate-chips-all	&	49	\\
    \multirow{40}{*}{~}	&	304	&	knife-handle	&	27	&	\multirow{17}{*}{~}	&	368	&	chocolate-cream-all	&	189	\\
    \multirow{40}{*}{~}	&	305	&	jam-knife-body	&	50	&	\multirow{17}{*}{~}	&	369	&	chocolate-cream-top	&	261	\\
    \multirow{40}{*}{~}	&	306	&	jam-knife-handle	&	53	&	\multirow{17}{*}{~}	&	370	&	chocolate-donut-all	&	35	\\
    \multirow{40}{*}{~}	&	307	&	metal-plate-edge	&	47	&	\multirow{17}{*}{~}	&	371	&	cookie-all	&	41	\\
    \multirow{40}{*}{~}	&	308	&	metal-spatula-body	&	71	&	\multirow{17}{*}{~}	&	372	&	cookie-end	&	57	\\
    \multirow{40}{*}{~}	&	309	&	metal-spatula-handle	&	123	&	\multirow{17}{*}{~}	&	373	&	cookie-head	&	1	\\
    \multirow{40}{*}{~}	&	310	&	pizza-spatula-body	&	70	&	\multirow{17}{*}{~}	&	374	&	cookies-all	&	49	\\
    \multirow{40}{*}{~}	&	311	&	pizza-spatula-handle	&	99	&	\multirow{17}{*}{~}	&	375	&	cookie-top	&	9	\\
    \multirow{40}{*}{~}	&	312	&	pizza-tray-edge	&	122	&	\multirow{17}{*}{~}	&	376	&	cream-all	&	139	\\
    \cline{5-8}
    \multirow{40}{*}{~}	&	313	&	plastic-spatula-body	&	328	&	\multirow{3}{*}{Drink}	&	377	&	tea-all	&	14	\\
    \multirow{40}{*}{~}	&	314	&	plastic-spatula-handle	&	531	&	\multirow{3}{*}{~}	&	378	&	tea-leaves-all	&	219	\\
    \multirow{40}{*}{~}	&	315	&	rolling-pin-body	&	198	&	\multirow{3}{*}{~}	&	379	&	whisk-head-top	&	9	\\
    \cline{5-8}
    \multirow{40}{*}{~}	&	316	&	rolling-pin-handle	&	174	&	\multirow{4}{*}{Uncategorised}	&	380	&	hand-left	&	23,305	\\
    \multirow{40}{*}{~}	&	317	&	rolling-pin-middle	&	82	&	\multirow{4}{*}{~}	&	381	&	hand-right	&	26,441	\\
    \multirow{40}{*}{~}	&	318	&	rolling-pin-miidle	&	5	&	\multirow{4}{*}{~}	&	382	&	towel-all	&	82	\\
    \multirow{40}{*}{~}	&	319	&	serrated-knife-body	&	38	&	\multirow{4}{*}{~}	&	383	&	towel-edge	&	38	\\
    \multirow{40}{*}{~}	&	320	&	serrated-knife-handle	&	43	&		&		&		&		\\
    \bottomrule
   \end{tabular}}
\end{table*}

\end{document}